%% file: main_tmc.tex
\documentclass[lettersize,journal]{IEEEtran}

\usepackage[ruled,linesnumbered]{algorithm2e}
\usepackage{subfigure}
\usepackage{multicol}
\usepackage{algorithmic}
\usepackage{caption}
\usepackage{bbding}
\usepackage{pifont}
\usepackage{graphicx} % 引入 graphicx 包
\usepackage{url}
\usepackage{gensymb}
\usepackage{hyperref}
\usepackage{amsmath}
\usepackage{booktabs}
\usepackage{xcolor}
\ifodd 1
\newcommand{\com}[1]{\textbf{\color{red}(COMMENT: #1)}} %comment of the 
\else

\newcommand{\com}[1]{}
\fi

\newcommand{\tmcrevise}[1]{{\color{black}{#1}}}

\def\fig{Fig.}

\def\eg{e.g.}

\ifCLASSINFOpdf
\else
\fi
\usepackage{balance}

\begin{document}

% \title{\fontsize{23.0pt}{\baselineskip}\selectfont {\revise{Transforming MAVs into Mobile Localization Infrastructures within Heterogeneous Swarms}}}
% \newcommand{\mycustomsize}{\fontsize{21.5}{\baselineskip}\selectfont}
% \title{\fontsize{22pt}{\baselineskip}\selectfont {\revise{Aerial Shepherds: Enabling Hierarchical Localization in Heterogeneous MAV Swarms}}}
\title{Enabling High-Frequency Cross-Modality Visual Positioning Service for Accurate Drone Landing}

% \title{\mycustomsize{TransformLoc: Transforming MAVs into Mobile Localization Infrastructures in Heterogeneous Swarms}}

% Aerial allies/shepherds: Enabling cooperative localization in Heterogeneous MAV Swarms

% Aerial allies/shepherds: Enabling hierarchical localization in Heterogeneous MAV Swarms

% Aerial allies/\textbf{shepherds}: Enabling hierarchical localization in Heterogeneous MAV Swarms


\author{
Haoyang Wang, 
Xinyu Luo,
Wenhua Ding,
Jingao Xu,
Xuecheng Chen,
% Zijian Xiao,
Ruiyang Duan,\\
Jialong Chen,
Haitao Zhang,
Yunhao Liu, ~\IEEEmembership{Fellow, ~IEEE},
Xinlei Chen, ~\IEEEmembership{Member, ~IEEE}

% \IEEEcompsocitemizethanks{
% \thanks{This paper was supported by the National Key R\&D program of China No. 2022YFC3300703, the Natural Science Foundation of China under Grant No. 62371269. Guangdong Innovative and Entrepreneurial Research Team Program No. 2021ZT09L197, Shenzhen 2022 Stabilization Support Program No. WDZC20220811103500001, and Tsinghua Shenzhen International Graduate School Cross-disciplinary Research and Innovation Fund Research Plan No. JC20220011. We acknowledge the support from the Tsinghua Shenzhen International Graduate School-Shenzhen Pengrui Endowed Professorship Scheme of Shenzhen Pengrui Foundation.}
% \thanks{This paper was supported by the Natural Science Foundation of China under Grant No. 62371269, Guangdong Innovative and Entrepreneurial Research Team Program No. 2021ZT09L197 and Meituan Academy of Robotics Shenzhen.}

% \thanks{A preliminary version of this article appeared in IEEE International Conference on Computer Communications (IEEE INFOCOM 2024).}

\thanks {Haoyang Wang, Xinyu Luo, Wenhua Ding, and Xuecheng Chen are with Shenzhen International Graduate School, Tsinghua University, China. 
E-mail: \{haoyang-22, luo-xy23, dingwh24, chenxc24\}@mails.tsinghua.edu.cn}

\thanks{Jingao Xu is with Carnegie Mellon University, USA. E-mail: jingaox@andrew.cmu.edu}

\thanks {Ruiyang Duan and Jialong Chen are with Meituan Academy of Robotics Shenzhen and Meituan Inc., China. (E-mail: \{duanruiyang, chenjialong02\}@meituan.com)}

\thanks{Haitao Zhang is with Shenzhen Institute for Advanced Study, University of Electronic Science and Technology of China, Shenzhen, China. (E-mail: zhanght23@uestc.edu.cn)}

\thanks {Yunhao Liu is with the School of Software and BNRist, Tsinghua University, Beijing 100084, China. 
Email: yunhaoliu@gmail.com}

% \IEEEcompsocthanksitem Zihong Lu is with Harbin Institute of Technology, China.\\
% E-mail: luzong2001@gmail.com

% \IEEEcompsocthanksitem  is with Shenzhen Key Laboratory of Ecological Remediation and Carbon Sequestration, Institute of Environment and Ecology, Tsinghua Shenzhen International Graduate School, Tsinghua University, Shenzhen, China.\\
% E-mail: hongcp@sz.tsinghua.edu.cn

% \thanks{ Chaopeng Hong and Xiao-Ping Zhang are with Shenzhen International Graduate School, Tsinghua University, China.
% E-mail: \{hongcp, xiaoping.zhang\}@sz.tsinghua.edu.cn}

\thanks{Xinlei Chen is with Shenzhen International Graduate School, Tsinghua University, China. 
E-mail: chen.xinlei@sz.tsinghua.edu.cn}

\thanks{Haoyang Wang and Xinyu Luo are co-primary authors.}
\thanks{ Corresponding author: Xinlei Chen.}
\thanks {Manuscript submitted Oct. 2025.}
% Major revision submitted December 2024. 

\vspace{-0.4cm}
}

% }


\markboth{IEEE TRANSACTIONS ON MOBILE COMPUTING}%
{Wang. H \MakeLowercase{\textit{et al.}}: EV-Pose}



\maketitle

\IEEEtitleabstractindextext{%

\begin{abstract}
After years of growth, drone-based delivery is transforming logistics.
\tmcrevise{At its core, real-time 6-DoF drone pose tracking enables precise flight control and accurate drone landing.}
% With the widespread accessibility of urban 3D maps, the Visual Positioning Service (VPS), a mobile pose estimation system, has been adapted to improve drone pose tracking in GPS-denied environments.
\tmcrevise{With the widespread availability of urban 3D maps, the Visual Positioning Service (VPS), a mobile pose estimation system, has been adapted to enhance drone pose tracking during the landing phase, as conventional systems like GPS are unreliable in urban environments due to signal attenuation and multi-path propagation.}
However, deploying the current VPS on drones faces limitations in both estimation accuracy and efficiency.
In this work, we redesign drone-oriented VPS with the event camera and introduce EV-Pose to enable accurate, high-frequency 6-DoF pose tracking for accurate drone landing.
EV-Pose introduces a spatio-temporal feature-instructed pose estimation module that extracts a temporal distance field to enable 3D point map matching for pose estimation;
and a motion-aware hierarchical fusion and optimization scheme to enhance the above estimation in accuracy and efficiency, by utilizing drone motion in the \textit{early stage} of event filtering and the \textit{later stage} of pose optimization.
Evaluation shows that EV-Pose achieves a rotation accuracy of 1.34$\degree$ and a translation accuracy of 6.9$mm$ with a tracking latency of 10.08$ms$, outperforming baselines by $>$50\%, \tmcrevise{thus enabling accurate drone landings.}
\textit{
% Artifacts will be released before publication.
Demo: \href{https://ev-pose.github.io/}{https://ev-pose.github.io/}.
}
\end{abstract}

}


% is not selected <OR> if conference mode is selected - because compsoc
% conference papers position the abstract like regular (non-compsoc)
\IEEEdisplaynontitleabstractindextext

% \fi
\IEEEpeerreviewmaketitle

% \IEEEraisesectionheading{\section{Introduction} \label{intro}}
% \input{intro_tmc}
% \input{systemoverview.tex}
% \input{systemdesign.tex}
% \input{evaluation.tex}
% \input{relatework.tex}
% \input{discussion.tex}
% \input{conclusion.tex}

\input{intro_mobicom}

% \input{intro_mobicom 0313}
\input{background}

\input{systemoverview_1209}

\input{systemdesign_1209}

\input{implementation.tex}
\input{evaluation.tex}
\input{relatework.tex}
% \input{discussion.tex}
\input{conclusion.tex}

% \newpage
% \input{Appendix.tex}

% % use section* for acknowledgment
% \ifCLASSOPTIONcompsoc
%   % The Computer Society usually uses the plural form
%   \section*{Acknowledgments}
% \else
%   % regular IEEE prefers the singular form
% \fi

% \input{Acknowledgement}

% \newpage

% \balance
\bibliographystyle{unsrt}
\bibliography{mobicom}

\end{document}

%% file: intro_mobicom.tex
% \vspace{-0.5cm}
\section{Introduction} \label{1}

\IEEEPARstart{D}{rone}-based delivery has been revolutionizing logistics by reducing delivery times, enhancing accessibility to congested areas in rush hours, and providing a sustainable alternative to conventional transportation~\cite{he2023acoustic, wang2022micnest, wang2025ultra, Global-Drones}.
% Central to the success of drone-based delivery is the accurate estimation of the drone's real-time six-degree-of-freedom (6-DoF) pose, which ensures precise navigation, effective obstacle avoidance, and reliable package drop-off~\cite{sharma2023beavis, qin2018vins, xu2022fast, seth2024aerobridge, ulrich2023real}.
% Central to the success of drone-based delivery is the \textit{landing phase}, where accurate estimation of the drone's real-time six-degree-of-freedom (6-DoF) pose is of paramount importance, since it ensures precise navigation, effective obstacle avoidance, and reliable package drop-off~\cite{sharma2023beavis, qin2018vins, xu2022fast, seth2024aerobridge, ulrich2023real}.
% The \textit{landing phase} is a critical determinant of the success of drone-based delivery, wherein accurate real-time six-degree-of-freedom (6-DoF) pose estimation is indispensable for achieving precise navigation, effective obstacle avoidance, and reliable package drop-off~\cite{sharma2023beavis, qin2018vins, xu2022fast, seth2024aerobridge, ulrich2023real}
\tmcrevise{
Central to success of drone-based delivery is the \textit{landing phase}, in which accurate real-time six-degree-of-freedom (6-DoF) pose estimation plays a pivotal role by enabling precise navigation, effective obstacle avoidance, and reliable package drop-off~\cite{sharma2023beavis, qin2018vins, xu2022fast, seth2024aerobridge}.
% However, conventional positioning systems such as GPS fail to achieve accurate pose estimation due to strong signal attenuation and multi-path propagation in urban environments during the \textit{landing phase}~\cite{Hu2024Square, gowda2016tracking, budiyono2012principles, groves2011shadow}.
However, conventional positioning systems such as GPS are inherently unreliable for accurate pose estimation during landing in urban areas, primarily due to severe signal attenuation and multi-path propagation effects~\cite{sun2023indoor, gowda2016tracking, budiyono2012principles}.
Without reliable systems to guarantee accurate and timely landings, operational efficiency will be compromised, and risks will be elevated in densely populated or high-traffic commercial zones \cite{albanese2021sardo}.
}

\begin{figure}[t]
    \setlength{\abovecaptionskip}{0.2cm} % height above Figure X caption
    \setlength{\belowcaptionskip}{0.01cm}
    \centering
        \includegraphics[width=1\columnwidth]{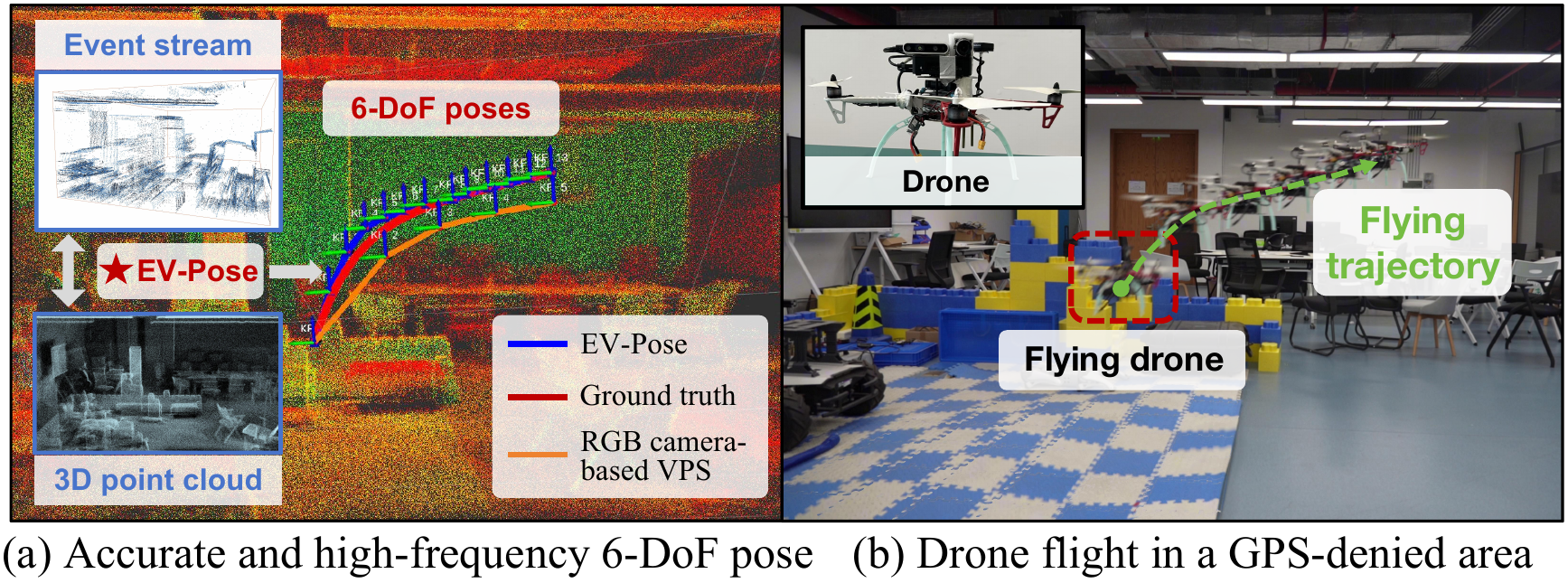}
        % \vspace{-0.2cm}
    \caption{
    EV-Pose estimates the drone's 6-DoF pose by redesigning the drone-oriented VPS with event cameras.
    As shown in (a), compared to conventional VPS systems, EV-Pose enables rapid and high-frequency drone pose tracking, ensuring precise flight control and landing as shown in (b).
    }
    \label{fig:intro}
    \vspace{-0.3cm}
\end{figure}

Recently, as urban 3D maps have become more widely accessible through platforms such as Google Earth, Mapbox, and Apple's Flyover, Visual Positioning Services (VPS) are increasingly being integrated into drones to achieve precise 6-DoF pose estimation \tmcrevise{during landing} in urban areas~\cite{pang2023ubipose, reinhardt2019using, hu2023efficient}. 
VPS uses video frames captured by an RGB camera. 
It extracts visual feature points from frames and matches them against a pre-built 3D environmental map (typically represented as a 3D point cloud) to accurately determine the drone's real-time 6-DoF pose \cite{cui2024vilam, recap, aerotraj, khan2024vrf, cip}.
It further improves the accuracy of these pose estimations by integrating data from onboard IMU sensors using Visual-Inertial Odometry (VIO) technology \cite{hu2023efficient, Zhou2022EDPLVO}.

While VPS shows great promise for accurate drone pose estimation \tmcrevise{during landing}, our field studies conducted in collaboration with a leading drone delivery company uncovers two major limitations when applying current VPS to drones:\\
\tmcrevise{
$(i)$ \textit{Motion blur negatively impacts the drone pose estimation accuracy}.
% Drone delivery systems are increasingly adopting higher speeds to shorten package delivery times.  
% For example, the DJI Matrice 4 can cruise at up to $v = 10$ m/s for urgent deliveries~\cite{DJI}.
Drone delivery systems need to rapidly and precisely adjust their pose during the landing process to ensure accurate touchdown at the designated location.
% However, the motion of the drone often causes video frames to suffer from severe motion blur
However, during flight, the drone’s motion causes video frames to be highly susceptible to severe motion blur, impairing feature point detection and pose estimation algorithms, ultimately resulting in increased errors in estimating drone pose \cite{argaw2021motion}.\\
}
% chang2021beyond
$(ii)$ \textit{The sampling rate mismatch between sensor modalities delays drone pose estimation.}
For precise flight control, drone pose estimation must operate in real-time (\eg,$>$100 Hz) to provide immediate feedback to the flight controller \cite{cheng2018end, sie2023batmobility}.
The onboard IMU typically operates at $<$200 Hz, whereas the camera captures frames at $<$30 Hz.
Accordingly, the camera's limited sampling rate constrains VIO pose updates to $<$30 Hz, forcing the controller to rely on IMU over extended periods, leading to accumulated errors and potential loss of control \cite{xu2022swarmmap, qin2018vins}.

Even worse, in practice, since urban maps are typically stored as 3D point clouds, the significant computational overhead required to match 2D images with 3D digital maps often limits drone pose estimation algorithms to operating at $<$1 Hz \cite{hu2023efficient, pang2023ubipose}.
All these issues necessitate a comprehensive rethinking of the visual positioning system \tmcrevise{designed for drone landing in urban areas}, spanning from hardware components to algorithmic designs. 
% All these issues necessitate a comprehensive rethinking of visual positioning system designs, spanning from hardware to algorithmic designs. 

This paper explores the feasibility of redesigning drone-oriented VPS with {\it event cameras}.
Event cameras are bio-inspired sensors that capture pixel-level intensity changes with a high dynamic range \cite{li2024taming, wang2025towards}.  
Compared to low frame-rate RGB cameras, they offer $ms$-level resolution and latency and thus can effectively capture scene dynamics without blurring~\cite{falanga2020dynamic, cao2024EventBoost}.  
As shown in \fig \ref{fig:intro}, dual advantage of absence of motion blurring and high temporal resolution makes event cameras a better alternative for addressing aforementioned limitations with current VPS systems.

Nevertheless, simply replacing RGB cameras with event cameras without adapting the current software stack and vision algorithms in VPS systems would not work due to the unique hardware characteristics of event cameras. We summarize two major challenges below.

\begin{figure}[t]
    \setlength{\abovecaptionskip}{0.15cm} % height above Figure X caption
    \setlength{\belowcaptionskip}{0.01cm}
    \centering
        \includegraphics[width=1\columnwidth]{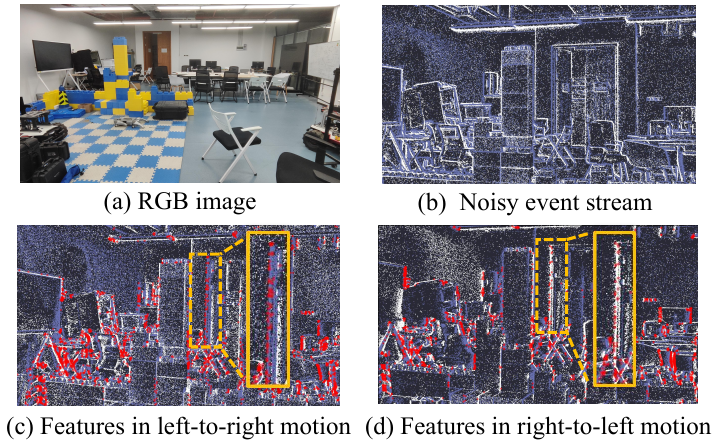}
        % \vspace{-0.2cm}
    \caption{
    Motivation study. 
    \textbf{(a)} RGB image. 
    \textbf{(b)} Noisy event stream. 
    \textbf{(c)} In left-to-right motion, event features extracted by Arc*, an event-based feature extraction algorithm \cite{alzugaray2022event}. 
    \textbf{(d)} Arc* features extracted during right-to-left motion.
    }
    \label{fig:motivation}
    \vspace{-0.3cm}
\end{figure} 

$\bullet$ \textbf{C1}: \textit{Extraction of repeatable vision features from the event stream for cross-modality vision feature matching.} 
Event cameras detect illumination changes at pixel-level granularity, enabling easy identification of object edges within a scene, as illustrated in \fig \ref{fig:motivation}(b).
However, unlike RGB images, where visual features typically remain consistent regardless of the camera's viewpoint, event cameras produce significantly different event patterns depending on the drone's direction of flight relative to the observed object.
Consequently, existing feature extraction algorithms, even those tailored to event cameras, face difficulties in consistently extracting repeatable features, as illustrated by \fig \ref{fig:motivation}(c) and \fig \ref{fig:motivation}(d).
This often results in numerous noisy points, negatively affecting the accuracy of matching visual features with the 3D point map.

$\bullet$ \textbf{C2}: \textit{Joint optimization of efficiency and accuracy in map matching-based pose estimation.}
Event cameras are highly sensitive to environmental changes, with even slight movements triggering numerous pixels and generating hundreds of events. 
As the drone moves, rapid scene changes lead to event bursts, generating thousands of event reports in a short time. 
Utilizing all the reported events for feature matching would introduce significant computational overhead, delaying drone pose estimations. 
Moreover, the inherent lack of semantic content and scale ambiguity in event cameras causes incorrect matching when encountering complex structures (\eg, repetitive textures), leading to an accuracy decline.

To conquer these challenges, we introduce \textbf{EV-Pose}, an \textbf{E}vent camera-enhanced \textbf{V}isual Positioning Service, which is a novel, accurate, and high-frequency event-based drone 6-DoF \textbf{Pose} tracking system (\fig \ref{fig:intro}).
EV-Pose is designed to function in urban canyon environments, where satellite-based systems lose accuracy, rendering them nearly useless.
With EV-Pose, drones can achieve 6-DoF pose tracking even in challenging conditions, ensuring efficient flight.

To address \textbf{C1}, we design a \textit{spatio-temporal feature instructed pose estimation (STPE)} module (§\ref{4}). 
Leveraging temporal relationships among events in an event stream, STPE first introduces a separate polarity time surface and extracts a temporal distance field as a feature representation. 
This feature enables cross-modality matching between 2D events and a 3D point map, aiding in global drone pose estimation, which can be treated as exteroceptive measurements.

To address \textbf{C2}, we propose a \textit{motion-aware hierarchical fusion and optimization (MHFO)} scheme (§\ref{5}) to enhance pose estimation in \textit{STPE}.
Event camera maintains temporal consistency with IMU, which provides drone motion information.
Incorporating motion information and structural data from a 3D point map, MHFO first predicts event polarity and performs fine-grained event filtering, fusing event camera with IMU at \textit{early stage} of raw data processing to improve estimation \textit{efficiency}.
Then, MHFO treats motion as proprioceptive measurements and integrates them with exteroceptive data using a factor graph, further fusing event camera and IMU at \textit{later stage} of pose optimization to improve \textit{accuracy}.

We implement EV-Pose and integrate it into ArduPilot, a widely used open-source drone flight controller, and conduct 20+ hours of field studies.
Our experiments also cover public datasets with challenging scene sequences.
The system is benchmarked against three SOTA pose tracking systems using translation error, rotation error, and tracking latency metrics.  
Evaluation results show that EV-Pose achieves an average rotation error of 1.34$\degree$ and a translation error of 6.9$mm$, with a tracking latency of 10.08$ms$, outperforming baselines by $>$ 50\%.
We also conduct extensive experiments in field studies to demonstrate robustness of EV-Pose, including various dynamic scenarios and illumination conditions.

% \noindent 
\textbf{Summary.} 
% In summary, 
The contributions of this work are as follows:\\
$\textbf{(1)}$ We propose \textit{EV-Pose}, as far as we are aware, \tmcrevise{the first system to redesign drone-oriented VPS with event cameras, enabling precise, high-frequency drone 6-DoF pose tracking for accurate drone flight control and landing.}\\
$\textbf{(2)}$ We design a \textit{STPE} module that leverages temporal relationships among events to extract a temporal distance field used in matching with 3D map for pose estimation, and a \textit{MHFO} scheme to enhance \textit{STPE} for accurate and efficient estimation by hierarchically utilizing drone motion in \textit{early stage} of event filtering and \textit{later stage} of pose optimization.\\
$\textbf{(3)}$ We fully implement EV-Pose and evaluate its performance through extensive field studies and experiments. Comparisons with SOTA systems demonstrate the significant advantages and application potential of EV-Pose.

%% file: background.tex
% \vspace{-0.3cm}
\section{Background and Motivation} \label{2}
\subsection{RGB Camera-based VPS}

\begin{figure}[t]
    \setlength{\abovecaptionskip}{0.15cm} % height above Figure X caption
    \setlength{\belowcaptionskip}{0.5cm}
    \centering
        \includegraphics[width=1\columnwidth]{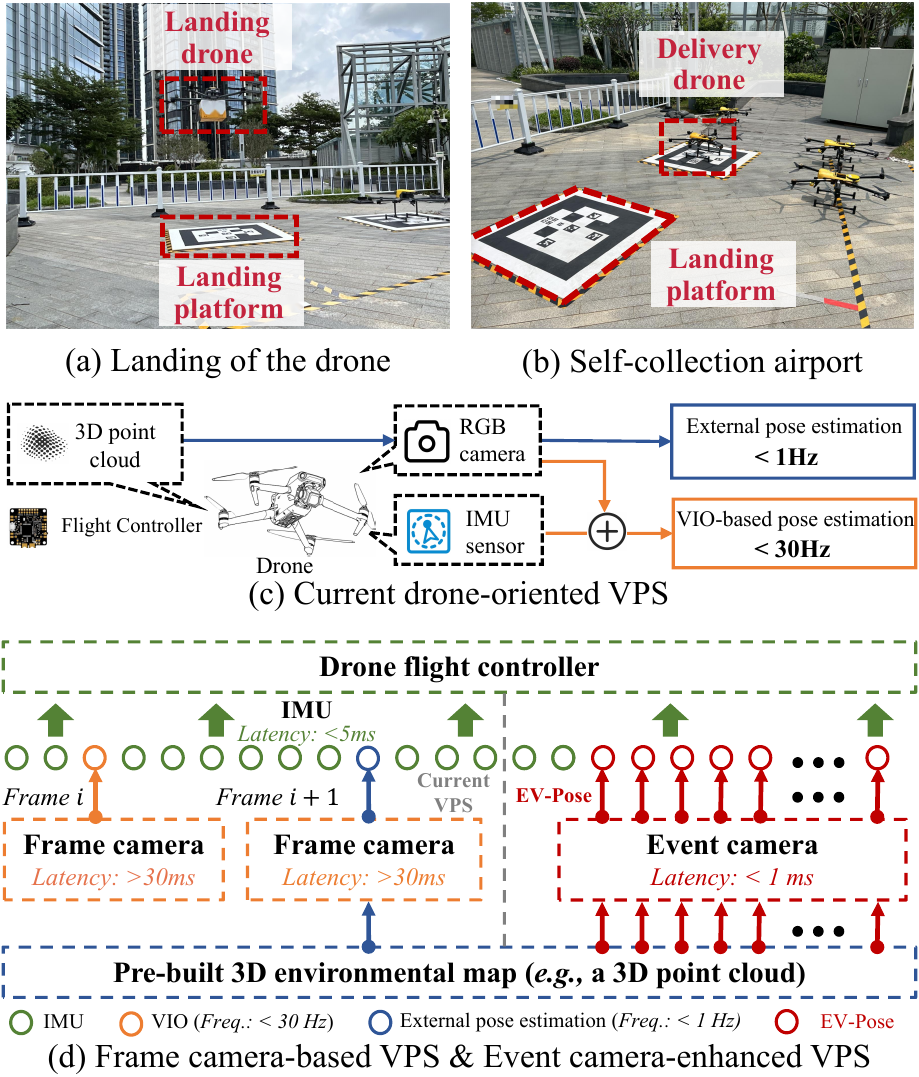}
        \caption{Current VPS and comparison of frame camera-based VPS \& event camera-enhanced VPS. 
        \textbf{(a)} Landing of drone.
        \textbf{(b)} Self-collection airport and landing platform.
        \textbf{(c)} Current VPS uses an RGB camera, an IMU, and 3D point clouds for pose estimation.
        \textbf{(d)} EV-Pose leverages event cameras for accurate and low-latency 6-DoF drone pose tracking.
        }
    \label{background}
    \vspace{-0.7cm}
\end{figure}

\textbf{Background.} 
Projected to reach a \$1 trillion market by 2040 \cite{Global-Drones}, the drone-driven low-altitude economy is revolutionizing industries with applications like on-demand delivery \cite{lee2021package}.  
At its core, 6-DoF drone pose tracking provides real-time drone location and orientation, supporting flight control, path planning, and landing, as shown in \fig \ref{background}(a) and \fig \ref{background}(b) \cite{li2024tmc_edgeslam2, cao2022edge}.  
With urban 3D maps widely available, VPS, a mobile pose estimation system leveraging environmental visual cues, has been adapted for drone 6-DoF tracking, particularly in GPS-denied environments \cite{reinhardt2019using, Hu2024Square}.  
Current VPS integrates an RGB camera, IMU, and pre-existing environmental data (\eg, a 3D point cloud) \cite{hu2023efficient, Zhou2022EDPLVO, ahmad2024cooperative}.   
As shown in \fig \ref{background}(c), the RGB camera and IMU facilitate VIO at $<$30 Hz, while the 3D point cloud supports external pose estimation at $<$1 Hz \cite{Hu20221D}.
Specifically, the external pose estimation is performed by extracting local features from a camera image and matching 2D descriptors to corresponding 3D map points. 
The prior map, typically pre-stored in the flight controller, is often generated from images or LiDAR data \cite{cui2024alphalidar, he2023vi, he2021vi}.

\textbf{Limitations of using current VPS on drone.} 
Despite the ability of VPS to provide accurate pose estimation for pedestrian navigation \cite{reinhardt2019using}, the high flight speed of drones presents significant challenges for frame camera-based VPS. 
Specifically, the motion blur in frame images disrupts feature point detection.  
As a result, pose estimation algorithms fail to establish correspondences between feature points and the map, degrading estimation accuracy.
At the same time, drone pose estimation must operate in real-time ($>$100 Hz) to provide immediate feedback to the flight controller for precise control.
However, despite onboard IMU running at $<$200 Hz, the long exposure time of frame cameras ($>$30 ms) limits VIO operating at $<$30 Hz; and the computational overhead of 2D-3D matching limits 3D point cloud-based global pose estimation to updates at $<$1 Hz. 
As shown in \fig \ref{background}(d), both VIO and map-based estimation rates are incompatible with the flight controller, forcing reliance on IMU and reducing accuracy.

\begin{figure}[t]
    \setlength{\abovecaptionskip}{0.15cm} % height above Figure X caption
    \setlength{\belowcaptionskip}{0.01cm}
    \centering
        \includegraphics[width=1\columnwidth]{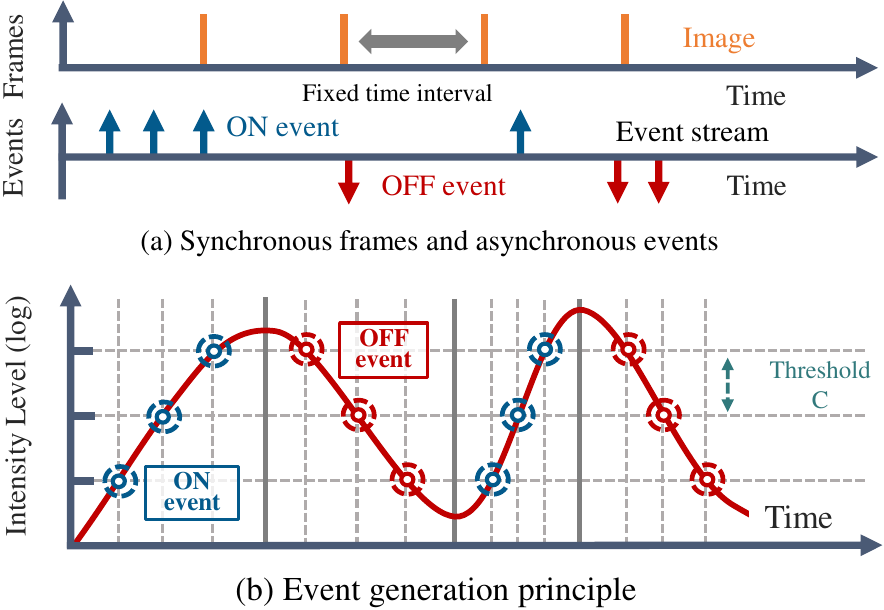}
        % \vspace{-0.2cm}
    \caption{
    Principles of frame cameras and event cameras. 
    \textbf{(a)} Frame camera uses a global shutter to capture synchronous frames. Each pixel of the event camera operates independently, generating events asynchronously.
    \textbf{(b)} Each pixel in event cameras generates events when intensity changes exceed a threshold: [ON] for increases, [OFF] for decreases.
    }
    \label{fig:primer}
    \vspace{-0.4cm}
\end{figure} 

\begin{figure*}[t]
    \setlength{\abovecaptionskip}{-0.2cm} % height above Figure X caption
    \centering
        \includegraphics[width=2.1\columnwidth]{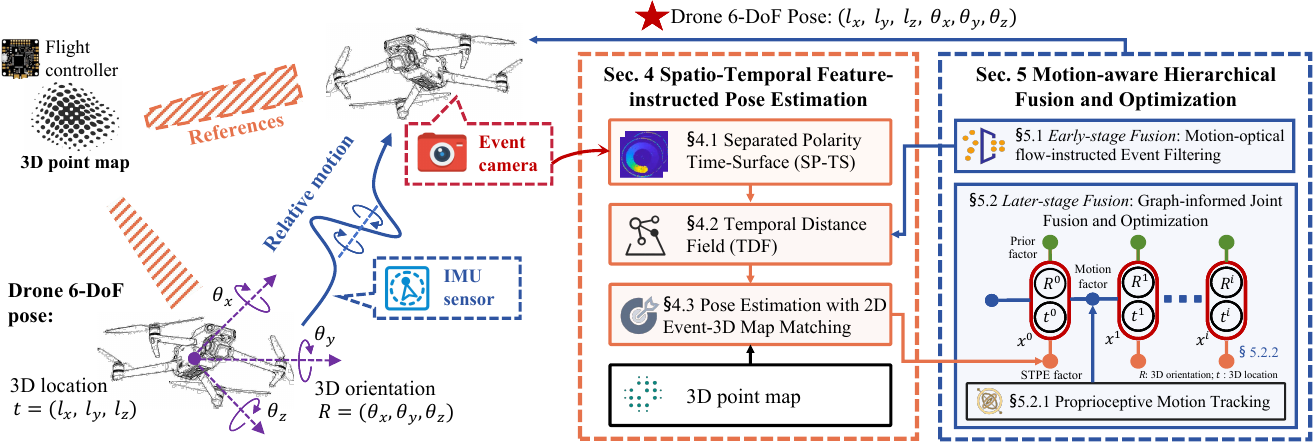}
        \vspace{0.1cm}
    \caption{System architecture of EV-Pose. }
    \label{overview}
    \vspace{-0.1cm}
\end{figure*}

% \vspace{-0.3cm}
% % \subsection{Event Camera-enhanced VPS}
\subsection{Event Camera-enhanced VPS}

\textbf{Principle of event camera.}
Event cameras are bio-inspired sensors that differ from traditional frame cameras \cite{gallego2020event}. 
Specifically, frame cameras capture synchronous images with a global shutter at fixed time intervals, while output of an event camera is event streams with $ms$-level resolution, as shown in \fig \ref{fig:primer}(a) \cite{he2024microsaccade}.
Each pixel in an event camera independently responds to changes in brightness asynchronously.
Each event \( e = (\boldsymbol{x}, i, p) \) represents a pixel at \(\boldsymbol{x} = (u, v)\) has undergone a predefined magnitude change in brightness at a time \( i \), as shown in \fig \ref{fig:primer}(b). 
\( p \) is polarity of intensity change, [ON] for brighter and [OFF] for darker.

\textbf{Advantages of event camera.}
The $ms$-level resolution and latency of event cameras make them highly suitable for capturing fast motion without motion blur, a common issue in frame cameras \cite{luo2024eventtracker}.
Additionally, due to their logarithmic intensity recording, event cameras achieve a 120 dB dynamic range, compared to 60 dB of frame cameras \cite{gallego2020event}, ensuring performance in extreme conditions (e.g., low light scenes).

\textbf{Redesign drone-oriented VPS with event camera.}
As shown in \fig \ref{background}(d), event cameras, with advantages in temporal resolution, hold great potential for revolutionizing drone-oriented VPS systems. 
3D point clouds are pre-stored in the flight controller. 
By redesigning the VPS with event cameras, we can leverage these prior 3D point clouds while ensuring temporal consistency with the IMU, enabling accurate and low-latency 6-DoF pose tracking for precise drone pose adjustment and landing \cite{wu2023flytracker, cao2024EventBoost}.
However, inherent characteristics of event cameras (\eg, motion-dependent, output of pixel-level intensity changes, and scale uncertainty) pose challenges for event-based tracking \cite{xu2023taming}. 
In this work, we present EV-Pose, which extracts spatio-temporal features from event streams and matches them against a 3D point map for pose estimation. 
Then, it enhances the accuracy and efficiency of the above estimation by mitigating event bursts and scale uncertainty through a hierarchical integration of motion information.

%% file: systemoverview_1209.tex
\vspace{0.3cm}
\section{Overview of EV-Pose}\label{3}

\subsection{Problem Statement}

\textbf{Variables description.}
The 6-DoF pose of drone includes its 3D location $(l_x, l_y, l_z)$ and 3D orientation $(\theta_x, \theta_y, \theta_z)$.
In EV-Pose, IMU is a standard sensor on a drone to provide motion information, and there are three coordinate systems: 
$(i)$ event camera coordinate system, $\mathtt{E}$; 
$(ii)$ IMU coordinate system, $\mathtt{M}$; 
and $(iii)$ 3D point map coordinate system, $\mathtt{P}$. 
The 3D point map is typically stored in the flight controller with known global coordinates.
Therefore, the 6-DoF pose of drone $x^i$ at each time $i$ can be treated as transformation from $\mathtt{E}$ to $\mathtt{P}$, described using a combination of rotation matrix ($R^i_{PE}$) and translation vector ($t^i_{PE}$).
The event camera and IMU are fixed on the drone, with relative pose remaining unchanged during movement, described as ($R^i_{EM}$, $t^i_{EM}$). 
Since global coordinates of 3D point map are known, ($R^i_{PE}$, $t^i_{PE}$) is equivalent to $(l_x, l_y, l_z, \theta_x, \theta_y, \theta_z)$. 
We use former for description, as it's widely used in flight control systems.

\textbf{Problem statement.}
EV-Pose leverages 3D point maps stored in flight controller, denoted as $\boldsymbol{\mathcal{P}} = \left\{ p^k \mid k \in \mathcal{N} \right\}$, and $\mathcal{N}$ is number of points. 
Subsequently, EV-Pose reads time-series signals from the event camera and IMU during a period $\mathcal{I}$, denoted as $\boldsymbol{\mathcal{E}} = \left\{ e^i \mid i \in \mathcal{I} \right\}$ and $\boldsymbol{\mathcal{M}} = \left\{ m^i \mid i \in \mathcal{I} \right\}$, respectively.
Using these three sets of data $\{\boldsymbol{\mathcal{P}}, \boldsymbol{\mathcal{E}}, \boldsymbol{\mathcal{M}}\}$ as input, EV-Pose calculates 6-DoF pose of drone $x^i = \left\{R^i_{IE}, t^i_{IE}\right\}$.

\begin{figure*}[t]
    \setlength{\abovecaptionskip}{0.2cm} % height above Figure X caption
    \setlength{\belowcaptionskip}{0.01cm}
    \centering
        \includegraphics[width=2.1\columnwidth]{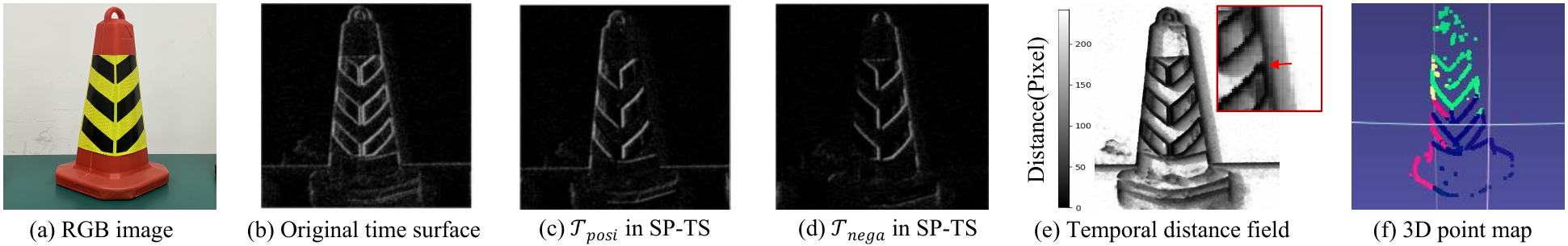}
        % \vspace{-0.2cm}
    \caption{ Illustration of SP-TS, temporal distance field, and 3D point map. 
    \textbf{(a)} Frame-based image; 
    \textbf{(b)} Original time surface; 
    \textbf{(c)} $\mathcal{T}_{posi}$ in SP-TS; 
    \textbf{(d)} $\mathcal{T}_{nega}$ in SP-TS; 
    \textbf{(e)} Temporal distance field; 
    \textbf{(f)} 3D sparse point reference map. 
    }
    \label{STPE}
    \vspace{-0.3cm}
\end{figure*} 

% \vspace{-0.3cm}
\subsection{Overview}
From a top-level perspective, we design EV-Pose, an event-based 6-DoF pose tracking system for drones that redesigns VPS with event cameras. 
EV-Pose leverages prior 3D point maps and temporal consistency between event camera and IMU to achieve accurate, efficient drone 6-DoF pose tracking.
As shown in \fig \ref{overview}, EV-Pose has two key parts:

% \noindent 
% \noindent 
$\bullet$ \textbf{Spatio-Temporal Feature-instructed Pose Estimation (STPE)} module (§\ref{4}).
This module first introduces the concept of a separated polarity time-surface, a novel spatio-temporal representation for event streams (§\ref{4.1}). 
Subsequently, it leverages the temporal relationships among events encoded in the time-surface to generate a distance field, which is then used as a feature representation for the event stream (§\ref{4.2}).
Finally, the 2D event-3D point map matching module models the drone pose estimation problem, which aligns the event stream’s distance field feature with the 3D point map, thus facilitating absolute pose estimation of the drone (§\ref{4.3}).

% \noindent 
% \noindent 
$\bullet$ \textbf{Motion-aware Hierarchical Fusion and Optimization (MHFO)} scheme (§\ref{5}).
This scheme first introduces motion-optical flow-instructed event filtering (§\ref{5.1}), which combines drone motion information with structural data from the 3D point map to predict event polarity and perform fine-grained event filtering. 
This approach fuses event camera with IMU at the \textit{early stage} of raw data processing, improving the \textit{efficiency} of matching-based pose estimation. 
This scheme then introduces a graph-informed joint fusion and optimization module (§\ref{5.3}),
This module first infers the drone’s relative motion through proprioceptive tracking and then uses a carefully designed factor graph to fuse these measurements with exteroceptive data from the STPF module. This fusion, performed at the \textit{later stage} of pose estimation, further improves the \textit{accuracy} of matching-based pose estimation.

\textbf{Relationship between STPE and MHFO.} 
STPE extracts a temporal distance field feature from the event stream and aligns it with a prior 3D point map to facilitate matching-based drone pose estimation.
To further enhance efficiency and accuracy of estimation, EV-Pose incorporates MHFO, which leverages drone motion information for early-stage event filtering, which reduces the number of events involved in matching, and later-stage pose optimization, which recovers scale and produces a 6-DoF trajectory with minimal drift.

%% file: systemdesign_1209.tex
% \vspace{-0.5cm}
\section{Spatio-Temporal Feature-instructed \\ Pose Estimation} \label{4}

\textbf{Challenge.} Event cameras produce asynchronous and noisy event streams, lacking inherent semantic information. 
As a result, it is challenging to extract meaningful and reliable features from the event stream and match them with the 3D point map for drone pose estimation.

\textbf{Observation.} 
Events output by an event camera exhibit unique spatio-temporal relationships, which can be considered a feature of the event stream. 
By extracting this feature and utilizing it for matching with the 3D point map, we can model the matching problem and estimate the drone's pose.

To realize this basic idea, we design a spatio-temporal feature-instructed pose estimation module to extract reliable features from the event stream and model the 2D event-3D point map matching problem for drone pose estimation.
This module first introduces a novel event representation, namely separated polarity time-surface (SP-TS) (§\ref{4.1}), mapping events according to polarity without sacrificing event features.
It then leverages spatio-temporal relationships encoded in SP-TS to extract a temporal distance field as a feature of event stream (§\ref{4.2}).
Finally, §\ref{4.3} employs a 3D point map as a reference, modeling pose estimation problem.

% \vspace{-0.3cm}
\subsection{Separated Polarity Time-Surface (SP-TS)} \label{4.1}

The high volume of noisy events makes it time-consuming to directly extract features from 3D meta event streams, as hundreds of events accumulate within a brief period.
Therefore, it is essential to develop a representation that can be used to effectively extract reliable features, enabling accurate pose estimation via 2D event-3D point map matching.

% \noindent 
\textbf{Consistency-based event filtering.} Event cameras are prone to noise from transistor circuits and other non-idealities, requiring pre-processing filtering at first. 
For the $k^{th}$ event $e^k$ with the timestamp $i$, we assess the timestamp of the most recent neighboring event in all directions ($i_{n(x)}$). 
Events with a time difference to the most recent neighboring event less than the threshold $T$ are retained, indicating object activity, while those exceeding $T$ are discarded as noise.

% \noindent 
\textbf{Spatio-temporal representation of events.}
We then introduce \textit{Separated Polarity Time-Surface (SP-TS)}, a lightweight event stream representation which preserves rich spatio-temporal information (\eg, edges, the most descriptive areas in the event stream). 
Time Surface (TS) is a 2D map in which each pixel value represents the timestamp of the most recent event at that location, and SP-TS maps events of two different polarities onto two separate 2D maps. 
In this way, the SP-TS emphasizes recent events rather than past ones. 
When using an exponential decay kernel, recent events are further highlighted, enhancing their significance.
Specifically, if $i_{last}$ is timestamp of the most recent event at each pixel $x = (u, v)^T$, then at time $i$ ($i \geq i_{last}(x)$), TS is defined as $\mathcal{T}(x, i) = exp \left(-\frac{i-i_{last}(x)}{\eta}\right)$, where $\eta$ denotes constant decay rate (\fig \ref{STPE}(b)).
For positive and negative events, SP-TS maintains separate TSs: $(i)$ $\mathcal{T}_{posi}$ is TS of positive events (\fig \ref{STPE}(c)), and $(ii)$ $\mathcal{T}_{nega}$ is TS of negative events (\fig \ref{STPE}(d)).
Constructing an SP-TS for each event stream requires \( O(N) \) time, as it involves only a single pass through the events.

% \noindent 
\textbf{Difference between SP-TS and original TS.} 
Original TS disregards event polarity when projecting event stream onto a 2D map, despite polarity encoding event features that can facilitate event filtering. 
In contrast, SP-TS retains polarity by maintaining separate TS for different polarities. 
As elaborated in §\ref{5.1}, SP-TS leverages polarity information in conjunction with the drone’s motion and environmental structure to refine event filtering, reducing the computational cost associated with 2D event-3D point map matching.

% \vspace{-0.3cm}
\subsection{Novel Event-based Feature: Temporal Distance Field } \label{4.2}
SP-TS captures the motion history of edges within a scene by tracking last motion timestamp at each pixel.
Generally, the values in TS $\mathcal{T}$ decrease gradually in one direction from these peak points, reflecting the edge's previous locations, while they decline sharply in the opposite direction, which can be seen as an anisotropic distance field.
By following direction where the values increase gradually, one can smoothly trace back to the edge's current location.

% \noindent 
\textbf{Temporal Distance Field (TDF) generation.}
We construct the TDF by negating and offsetting $\mathcal{T}$: $\overline{\mathcal{T}} = 1 - \mathcal{T}$. 
In the TDF, the pixel values at the edges are smaller, and the direction in which the values increase corresponds to the distance from the edge, creating a distance field (as shown in \fig \ref{STPE}(e)). 
The TDF can then be used as a perspective feature to evaluate the reprojected 3d points in the 2D event-3D map matching phase. 

% \noindent 
\textbf{Benefits of TDF.} In this way, the 2D event-3D point map matching can be formulated as a minimization problem. The values in the TDF serve as residuals, and the TS gradient provides the direction for optimization.

\begin{figure}[t]
    \setlength{\abovecaptionskip}{0.15cm} % height above Figure X caption
    \setlength{\belowcaptionskip}{0.01cm}
    \centering
        \includegraphics[width=1\columnwidth]{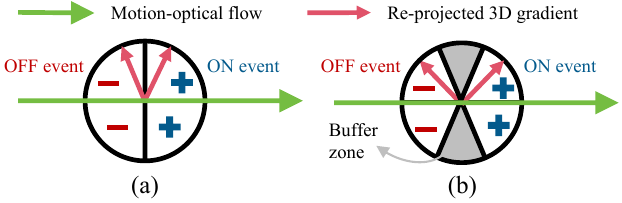}
    \caption{
    Illustration of event filtering. 
    \textbf{(a)} Prediction of the polarity using re-projected 3D gradient and optical flow; 
    \textbf{(b)} The buffer zone used in the prediction.
    If the angle is in the buffer zone, all events are used in point patch.
    }
    \label{optical}
    \vspace{-0.3cm}
\end{figure} 

% \vspace{-0.3cm}
\subsection{Pose Estimation with 2D Event-3D Map Matching} \label{4.3}
After applying the TDF as event perspective feature, we design an Event-Map Matching module to match TDF $\overline{\mathcal{T}}$ with the point cloud $\boldsymbol{\mathcal{P}}$ from the 3D point map (\fig \ref{STPE}(f)), outputting the preliminary estimated pose of the drone, $\hat{x}^i$. 

% \noindent
\textbf{Problem formulation of pose estimation.} 
This matching module first utilizes the candidate pose of the drone to wrap the point cloud onto the 2D plane of the event camera. 
It then matches the warping results with local minima of $\overline{\mathcal{T}}$ and adjusts the candidate pose based on the alignment results:
% \vspace{-0.2cm}
\begin{equation}
\hat{x}^i = \underset{R^i_{PE}, t^i_{PE}}{\arg \min} \sum_{p \in \boldsymbol{\mathcal{P}}} \rho\left(\overline{\mathcal{T}}(\pi (W(p; R_{PE}^i, t_{PE}^i)))^2\right),
% \vspace{-0.2cm}
\end{equation}
where $\rho$ is the robust loss function, $\pi$ is the projection function of the event camera, and $W(p; R_{PE}^i, t_{PE}^i) = R_{PE}^i \cdot p + t_{PE}^i$ utilizes the relative pose of the drone to transform $p \in \boldsymbol{\mathcal{P}}$.
% , $x^i = \left\{R^i_{IE}, t^i_{IE}\right\}$,
By solving this problem with a least-squares solver \cite{Agarwal_Ceres_Solver_2022}, we can estimate preliminary absolute pose of the drone.

% \noindent
\textbf{Latency and error source analysis.} 
Theoretically, events are triggered asynchronously with $ms$-level resolution, allowing high-frequency pose updates. 
However, incorporating all noisy events in 2D event–3D point map matching significantly increases computational overhead, reducing estimation efficiency. 
Additionally, the lack of semantic information and scale ambiguity in event streams causes mismatches in complex structures, leading to decreased accuracy.
Therefore, we further incorporate the drone’s motion and environmental structure to enhance estimation efficiency and accuracy.

\vspace{0.3cm}
\section{Motion-aware Hierarchical Fusion and Optimization} \label{5}

% \noindent 
\textbf{Challenge.}
Event cameras are highly sensitive to environmental changes. As drone moves, rapid scene variations generate thousands of event reports in a short time. 
Using all events for matching reduces estimation efficiency. 
A lack of semantic content and scale ambiguity in event cameras also leads to incorrect matching, causing accuracy decline.

% \noindent 
\textbf{Observation.}
Our design is based on two observations: 
$(i)$ By leveraging time-series motion and structural information in the 3D point map, we can predict the polarity of upcoming events and filter out irrelevant ones. 
$(ii)$ The pose estimation from the STPE module provides \textit{exteroceptive} measurements, while the drone motion offers \textit{proprioceptive} measurements. These two independent yet complementary ways can be jointly fused and optimized to calibrate each other, producing a 6-DoF trajectory with minimal drift.

% \vspace{-0.3cm}
\begin{algorithm}[t]
\caption{Pose estimation via 2D event–3D map matching, enhanced by event filtering}
\setlength{\abovedisplayskip}{3pt}
\setlength{\belowdisplayskip}{-1cm}
\label{argSTPE}
\KwData{Event stream $\boldsymbol{\mathcal{E}}$; Motion information $m^i$; Prior 3D point cloud  $\boldsymbol{\mathcal{P}}$. }
\KwResult{Pose estimation of the drone $x^i = \{R^i_{IE}, t^i_{IE}\}$.}

\textit{\% Event representation and processing.}\\
Represent $\boldsymbol{\mathcal{E}}$ as SP-TS, $\{\mathcal{T}_{posi}, \mathcal{T}_{nega}\}$ (\underline{§ \ref{4.1}});\\
Extract TDF of $\{\mathcal{T}_{posi}, \mathcal{T}_{nega}\}$: $\overline{\mathcal{T}}_{posi}$ and $\overline{\mathcal{T}}_{nege}$ (\underline{§ \ref{4.2}});\\
\For{iteration $i$ in $N$}
{
\textit{\textbf{\% Early-stage Fusion}: Motion-optical flow-instructed Event Filtering.}\\
Project $\boldsymbol{\mathcal{P}}$ into 2D plane of event camera using $\hat{x}^i$;\\
\For{each point $p^k$ in $\boldsymbol{\mathcal{P}}$}
{
    Predict the polarity of events in patch of \( p^k \) using \( m^i \), then apply it for event filtering, yielding \( \overline{\mathcal{T}}_{align} \) (\underline{§ \ref{5.1}}).
}
\textit{\% Efficient 2D event-3D point map matching.}\\
Conduct 2D event-3D point map matching using $\boldsymbol{\mathcal{P}}$ and TDF of $\overline{\mathcal{T}}_{align}$, then update $\hat{x}^i$ (\underline{§ \ref{4.3}}).
}

\end{algorithm}

\begin{figure*}[t]
    \setlength{\abovecaptionskip}{0.2cm} % height above Figure X caption
    \centering
        \includegraphics[width=2.1\columnwidth]{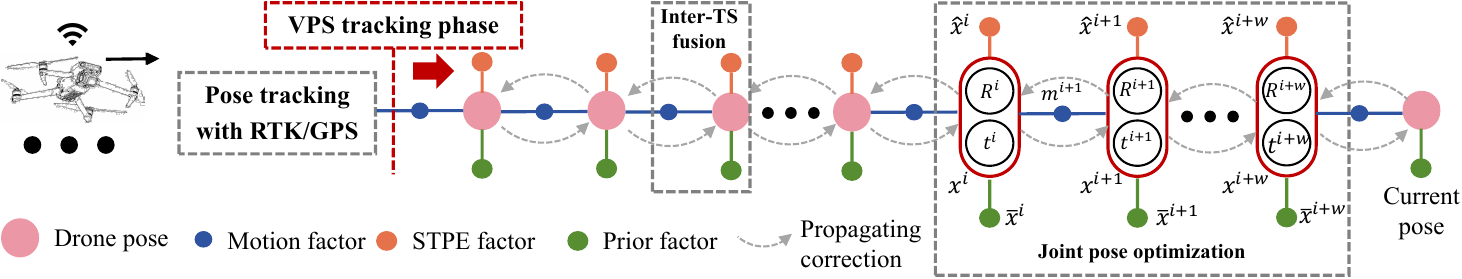}
        % \vspace{0.0cm}
    \caption{Illustration of the graph-informed joint fusion and optimization.
    }
    \label{fusion}
    \vspace{-0.25cm}
\end{figure*} 

To implement this concept, we propose a motion-aware hierarchical fusion and optimization scheme. 
The scheme starts with motion–optical flow–guided event filtering (§\ref{5.1}), which combines drone motion data and structural information from a 3D point map to predict event polarity and perform precise filtering. 
By fusing an event camera with IMU at the \textit{early stage} of raw data processing, it improves \textit{efficiency} of pose estimation based on 2D event–3D map matching.
Next, the scheme introduces a graph-informed joint fusion and optimization module (§\ref{5.3}). This module first infers the drone’s relative motion using proprioceptive motion tracking, then employs a carefully designed factor graph to fuse these proprioceptive measurements with exteroceptive data from the STPF module. This joint fusion further refines the pose estimation in the \textit{later stage}, improving the \textit{accuracy} of 2D event–3D map matching–based pose estimation.

% \vspace{-0.4cm}
\subsection{\textbf{Early-stage Fusion}: \\Motion-optical flow-instructed Event Filtering} \label{5.1}
In this part, we fuse the event camera and IMU at the early stage of raw data processing to enhance the efficiency of 2D event–3D map matching–based pose estimation.
As shown in Line 1–2 of Alg.\ref{argSTPE}, we first calculate the SP-TS, resulting in $\mathcal{T}_{posi}$ and $\mathcal{T}_{nega}$. 
Subsequently, by negating and offsetting $\mathcal{T}_{posi}$ and $\mathcal{T}_{nega}$, we obtain the TDFs: $\overline{\mathcal{T}}_{posi}$ and $\overline{\mathcal{T}}_{nega}$.

% \noindent
\textbf{Optical flow-instructed event filtering} (Line 5–9 of Alg.\ref{argSTPE}).
To avoid deviations in 2D event-3D point map matching caused by noise in event data, we first project the 3D gradients (\eg, normal vector) of the 3D point map onto the 2D plane of the event camera using the candidate pose of the drone. 
We then estimate the optical flow by combining point depth and motion signals from IMU \cite{beauchemin1995computation}.
Subsequently, we predict the polarity of the events triggered by the points based on the relationship between the 3D gradient of points and optical flow.
Specifically, as illustrated in \fig \ref{optical}(a), if the angle between the re-projected 3D gradient and the optical flow is less than 90$\degree$, we predict a positive event; otherwise, we predict a negative event. 

% \noindent
\textbf{Adaptive buffer zone}.
There may be errors in the optical flow and re-projected 3D gradients. 
Therefore, we introduce a buffer zone in the prediction process as shown in \fig \ref{optical}(b).
If the angle between the re-projected 3D gradient and the optical flow falls within the buffer zone, we consider the predicted event polarity to be unreliable and use events from both $\mathcal{T}_{posi}$ and $\mathcal{T}_{nega}$ in the corresponding patch of the point for matching. 
Otherwise, we use events from the TS corresponding to the predicted polarity in this patch.

% \noindent
\textbf{Enhance efficiency of matching-based pose estimation} (Line 10–11 of Alg.\ref{argSTPE}).
This process generates $\overline{\mathcal{T}}_{align}$ and corresponding TDF, filtering redundant events and reducing the number of events for 2D event-3D map matching, thereby improving drone pose estimation efficiency in §\ref{4.3}.

\subsection{\textbf{Later-stage Fusion}: \\Graph-informed Joint Fusion and Optimization} \label{5.3}

\subsubsection{\textbf{Proprioceptive motion tracking}} 
The measurements from the IMU provide relative motion information of the drone, including those from the accelerometer and gyroscope in $\mathtt{M}$. 
By using the pose relationship between $\mathtt{M}$ and $\mathtt{E}$, we transform these measurements into $\mathtt{E}$. 
The raw data from the accelerometer represents the sum of the gravitational force and the forces providing acceleration to the drone, while the gyroscope measurements reflect the angular velocity of the drone. 
We denote the raw accelerometer and gyroscope measurements at time $i$ as $\hat{a}^i$ and $\hat{\omega}^i$, respectively. 
$\hat{a}^i=a^i+g+n_a + b_a$ and $\hat{\omega}^i=\omega^i+n_\omega + b_\omega$, where $a^i$ and $\omega^i$ are the real acceleration and angular velocity, $g$ is gravity, and $n_a$, $n_\omega$ are Gaussian measurement noises, $b_a$, $b_\omega$ are bias of the accelerometer and gyroscope.

Taking $\mathcal{T}_{posi}$ as an example, by performing pre-integration of the IMU time-series signals, we can effectively summarize the changes in the drone's pose between $\mathcal{T}_{posi}^i$ and $\mathcal{T}_{posi}^{i+1}$.
In this work, we incorporate the continuous-time quaternion-based derivation of the IMU pre-integration approach \cite{qin2018vins}. 
Through this process, we calculate relative motion $m = \left \{ \Delta x^i, \Delta v^i\right\}$, where $\Delta x^i = \{\Delta t^i, \Delta R^i\}$. 

\subsubsection{\textbf{Joint fusion and optimization}} 
By utilizing the exteroceptive absolute pose provided by the STPE module and the proprioceptive relative motion calculated from IMU signal, along with prior knowledge of the drone (\eg, drone flight characteristics), we have meticulously designed a factor graph-based joint fusion and optimization approach, which integrates various independent yet complementary factors to enhance accuracy of drone pose estimation \cite{thrun2002probabilistic}. 

\noindent 
$\bullet$ \textbf{Prior factor.}
Prior refers to the probability distribution of the drone's pose at time $i$, $p(x^i)$, in the absence of any measurements. 
We employ a constant velocity model, which assumes that the drone moves and rotates at a constant velocity over short periods—a widely used assumption in SLAM—to derive this prior. 
Therefore, the drone's prior $\bar{x}^i = \{R_{IE}^i, t_{IE}^i\}$ is described by $\bar{R}_{IE}^i = R_{IE}^{i - 1} (R_{IE}^{i - 2})^T R_{IE}^{i - 1}$ and $\bar{t}_{IE}^i = 2t_{IE}^{i - 1} - t_{IE}^{i - 2}$, with $p(x^i \mid x^{i - 1}, x^{i - 2}) \sim \mathcal{N}(\bar{x}^i, \sigma_x)$ and the prior factor is: 
% described as:
\begin{equation}
    E^i_{Prior} = -\log p(x^i) \propto \|x^i - \bar{x}^i\|_{\sigma_x}.
\end{equation}

\noindent 
$\bullet$ \textbf{STPE factor: \textit{Exteroceptive} measurements.}
The likelihood of the STPE module, $p(\hat{x}^i \mid x^i)$ refers to the probability distribution of the estimated pose $\hat{x}^i$ given $x^i$, where $\hat{x}^i$ provides the exteroceptive absolute pose. 
The estimated pose is affected by Gaussian noise with a standard deviation of $\sigma_{\hat{x}^i}$.
We define the STPE factor as:
\begin{equation}
\begin{aligned}
    E^i_{STPE} = -\log p(\hat{x}^i \mid x^i) \propto \|x^{i} -\hat{x}^i\|^2_{\sigma_{\hat{x}^i}}.
\end{aligned}
\end{equation}

\noindent 
$\bullet$ \textbf{Motion factor: \textit{Proprioceptive} measurements.}
The measurement likelihood of motion tracking $p(\Delta x^i \mid x^i, x^{i + 1})$ indicates the distribution of the measured motion distance and angle at a given pose, providing proprioceptive relative motion.
We define the IMU motion factor as:
\begin{equation}
    E^i_{Motion} = -\log p(\Delta x^i \mid x^i, x^{i + 1}) \propto \|x^{i + 1} - x^{i} - \Delta x^i\|^2_{\sigma_{M}}.
\end{equation}
The pre-integration of IMU is affected by Gaussian noise with a standard deviation of $\sigma_{M}$.

\fig \ref{fusion} illustrates the construction and optimization of the factor graph. We aim to optimize the value of nodes (poses) based on all factors, and this process has two stages:

\noindent 
$\bullet$ \textbf{Inter-TS fusion: Put together.}
When receiving preliminary absolute pose estimation from the STPE module, Inter-TS fusion is executed to provide an immediate pose optimization result. 
Specifically, Inter-TS fusion problem is modeled as:
\begin{equation}
\begin{aligned}
    \hat{x}^i &= \underset{x^i}{\arg \min} \left(E^i_{Prior} + E^i_{STPE} + E^i_{Motion}\right), \\
    x^i &= \left\{R^i_{IE}, t^i_{IE}\right\},
\end{aligned}
\end{equation}
where $E^i_{Prior}$ is the prior factor, $E^i_{STPE}$ is the STPE factor, functioning as an exteroceptive measurement, and $E^i_{Motion}$ is the motion factor, serving as a proprioceptive measurement.

\noindent
$\bullet$ \textbf{Joint pose optimization: Accuracy enhancement.}
Every few poses, joint pose optimization is activated to recover the scale and correct accumulated errors, enhancing the accuracy of 2D event-3D map matching.
This module is based on a sliding window and aims to jointly optimize all poses within the window. 
Denoting sliding window as $\mathcal{W}$, the optimization problem is modeled as:
\begin{equation}
\begin{aligned}
    \hat{\mathcal{X}} &= \underset{\mathcal{X}} {\arg \min} \sum_{i \in \mathcal{W}} \left(E^i_{Prior} + E^i_{STPE} + E^i_{IMU}\right), \\
    \mathcal{X} &= \underset{i \in \mathcal{W}} {\bigcup} \left\{R^i_{IE}, t^i_{IE}\right\}.
\end{aligned}
\end{equation}
These two problems can be efficiently solved using the incremental factor graph optimization method, which iteratively corrects each node value based on connected factors, resulting in the corrected trajectory.

%% file: implementation.tex
\section{Implementation}

\subsection{Testbed Configuration.}
\begin{figure}[t]
    \setlength{\abovecaptionskip}{0.2cm} % height above Figure X caption
    \setlength{\belowcaptionskip}{0.02cm}
    \centering
        \includegraphics[width=1\columnwidth]{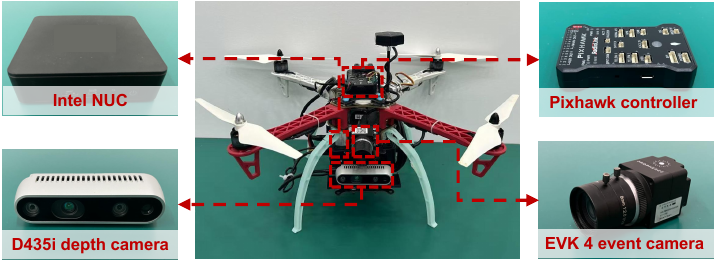}
    \caption{Testbed configuration of EV-Pose platform.}
    \label{implement}
    \vspace{-0.3cm}
\end{figure}

As illustrated in \fig \ref{implement}, EV-Pose is implemented using a 450 mm-wide drone equipped with $(i)$ a Prophesee EVK4 HD event camera with 1280 × 720 resolution; $(ii)$ a D435i depth camera for frame images capture; and $(iii)$ a Pixhawk flight controller for drone control and IMU measurements. 
EV-Pose runs on an Intel NUC with a Core i7 CPU, 16GB RAM, and Ubuntu 20.04.
Indoor and outdoor environment mapping is completed in advance using the Livos MID-360 LiDAR and FAST-LIO2 algorithm \cite{xu2022fast}.
All EV-Pose algorithms are implemented in C++ and ROS \cite{ROS}.

% \vspace{-0.3cm}
\subsection{Map Construction and Optimization.}
To better match the point cloud and event feature, we introduce a map construction method that emphasizes object contours over textures by fitting planes to raw point cloud data and extracting contours from these planes.
Specifically, point cloud data, represented as ${p_i = (x_i, y_i, z_i)}$, is processed using Principal Component Analysis (PCA) to compute normals and curvatures. 
For each point $p_i$, its $k$-nearest neighbors ${p_j}$ define the covariance matrix, 
where eigenvalues $\lambda_0 \leq \lambda_1 \leq \lambda_2$ and eigenvectors $v_0, v_1, v_2$ are obtained via singular value decomposition. The normal is given by $v_0$ and the curvature at $p_i$ is: $\sigma(p_i) = \lambda_0 / (\lambda_0 + \lambda_1 + \lambda_2)$.
Plane fitting is performed using a region-growing algorithm that groups points based on the angular difference between their normals. Smoothness constraints identify clusters, forming segmented planes. A contour-tracking algorithm then extracts closed boundary contours from the segmented regions.

\begin{figure*}[t]
    \setlength{\abovecaptionskip}{0.01cm} % height above Figure X caption
    \setlength{\belowcaptionskip}{0.01cm}
    \centering
        \includegraphics[width=2\columnwidth]{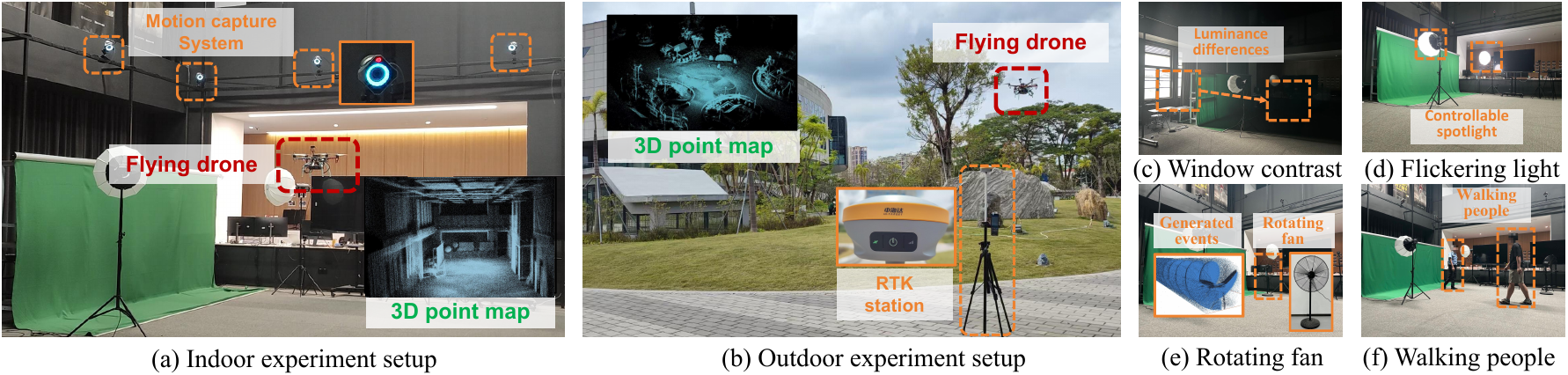}
        \vspace{0.1cm}
    \caption{Experimental scenarios of EV-Pose.
    }
    \label{setup}
    \vspace{-0.55cm}
\end{figure*}

\begin{figure*}
\setlength{\abovecaptionskip}{-0.05cm} 
\setlength{\belowcaptionskip}{-0.35cm}
\setlength{\subfigcapskip}{-0.2cm}
    \begin{minipage}[b]{1.03\columnwidth}
        \subfigure[CDF of ATE]{
            \centering
            \includegraphics[width=0.466\columnwidth]{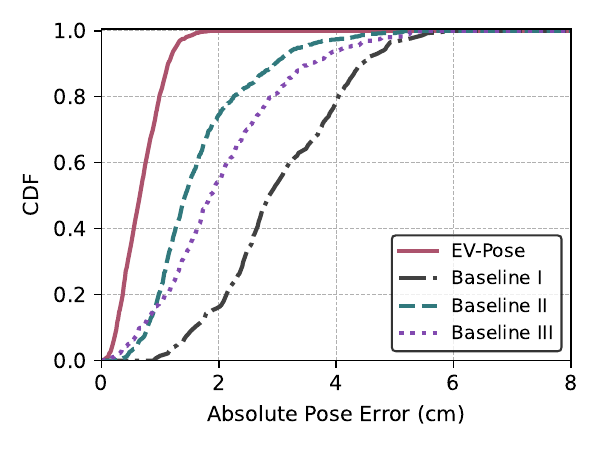}
        }
        \subfigure[CDF of ARE]{
            \centering
            \includegraphics[width=0.466\columnwidth]{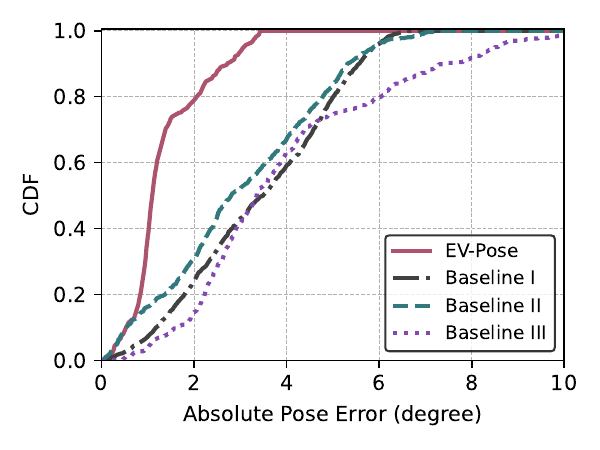}
        }
        \caption{Real-world indoor experiments.}
        \label{fig:overall_indoor}
    \end{minipage}
    \begin{minipage}[b]{1.03\columnwidth}
        \subfigure[CDF of ATE]{
            \centering
            \includegraphics[width=0.466\columnwidth]{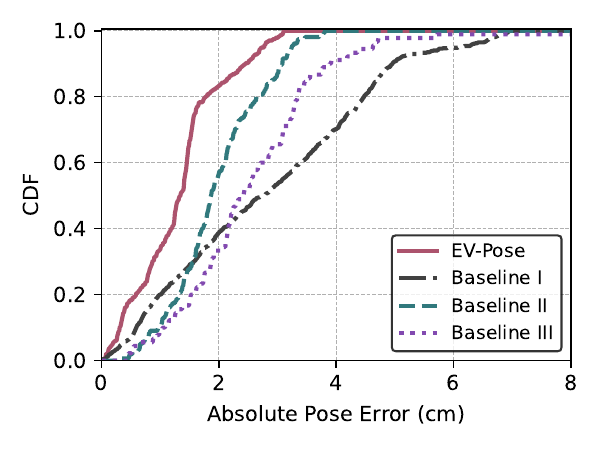}
        }
        \subfigure[CDF of ARE]{
            \centering
            \includegraphics[width=0.466\columnwidth]{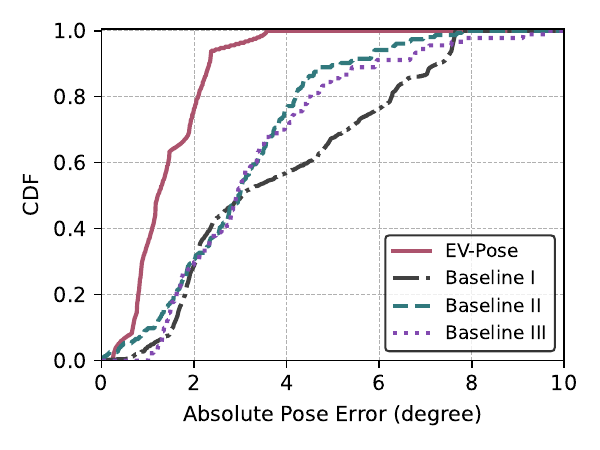}
        }
        \caption{Real-world outdoor experiments.}
        \label{fig:overall_outdoor}
    \end{minipage}
% \vspace{-0.1cm}
\end{figure*}

% \vspace{-0.2cm}
\subsection{Push Limit of Accuracy and Efficiency}

% \noindent
\textbf{Ego-motion instructed infrastructure extraction.} 
Moving objects within the field of view (FoV) often disrupt alignment and need to be filtered.
We propose a motion analysis method to identify and prioritize stationary objects, which provide reliable edge features for localization. 
% Building on this, we locate reference infrastructures. 
For a sequence of events \(\mathcal{E}\) and IMU data \(\mathcal{M}\) over an interval \([0, \delta i]\), each event \(e = (\boldsymbol{x}, i, p)\), where \(\boldsymbol{x}\) and \(p\) represent pixel coordinates and polarity, is analyzed.
The predicted location \(\hat{\boldsymbol{x}}_{\delta i}\) is derived by integrating IMU data, followed by coordinate transformations. 
If \(e_{\delta i}\) is triggered by stationary objects, predicted location \(\hat{\boldsymbol{x}}_{\delta i}\) should align with actual location \(\boldsymbol{x}_{\delta i}\), as changes in stationary objects are solely due to drone motion.

\textbf{Perspective movement in point cloud.} 
In Sec. 4, we detailed our 2D event-3D point map algorithm.
To further improve pose estimation efficiency, once the initial pose is established, we restrict the point cloud range using the drone's current pose and FoV.
Only points within the FoV are used for alignment. 
As the drone moves, the perspective dynamically shifts to include new regions of the point cloud for pose estimation.
To accommodate potential errors in pose estimation, we expand the perspective range.
Also, we integrate occlusion handling by prioritizing nearby point clouds to the estimated pose, improving matching efficiency. 

%% file: evaluation.tex
% \vspace{-0.15cm}
\section{Evaluation}\label{6}
\subsection{Experiment Methodology}

% \noindent
\textbf{Field studies.} 
We conduct field studies of EV-Pose both indoors and outdoors to evaluate its pose tracking performance, as shown in \fig \ref{setup}.
The drone flies along a square spiral trajectory within the test field.
The environment is pre-mapped, and the drone operates in the 3D point-mapped environment for pose estimation. 
In indoor settings, the drone's pose ground truth is obtained using a CHINGMU Motion Capture system with 16 MC1300 infrared cameras operating at 210 FPS \cite{motioncapture}. 
For outdoor settings, we deployed a private RTK station using a Hi-Target D8 to provide accurate ground truth \cite{RTK}.
We conduct 20+ hours of extensive experiments, collecting 200+GB of raw data.

% \noindent 
% \vspace{-0.3cm}
\textbf{Dataset-based studies.} 
We also perform extensive experiments on public datasets to evaluate the performance of EV-Pose thoroughly. Specifically, we test on two datasets to demonstrate its effectiveness across different scenarios:
$(i)$ the TUM-VIE dataset (denoted as T) \cite{klenk2021tumvie},
$(ii)$ the VECtor dataset (denoted as V) \cite{gao2022vector}.
The 3D point reference maps for both datasets are provided by depth cameras.

% \noindent 
\textbf{Comparative methods.}
To comprehensively evaluate the performance of EV-Pose, which is the first \textit{event camera-enhanced VPS}, we compare it with three related systems: \\
% ORB-SLAM3
\textit{\textbf{(i)} Baseline I \cite{guan2023pl}: A SOTA event camera-based VIO}, which uses event streams and IMU data for drone pose estimation (the event–IMU version of \cite{guan2023pl}), validating the benefit of incorporating prior 3D point maps in EV-Pose.
Since \cite{guan2023pl} does not provide publicly available code, for a fair comparison, we implement it based on \cite{vidal2018ultimate} 
by adding line-based event features to the feature set used in VIO-based pose estimation.\\
\textit{\textbf{(ii) }Baseline II \cite{hu2023efficient}: A SOTA frame camera-based VPS}. 
This method first extracts point and line features from frame images and integrates IMU data for VIO. 
By leveraging prior 3D point maps, it then establishes 2D-3D correspondences through feature matching between 2D images and 3D map points, enabling drone pose estimation.
This method is used to evaluate the effectiveness of EV-Pose in redesigning drone-oriented VPS with event cameras.
Since \cite{hu2023efficient} does not provide publicly available code, for a fair comparison, we implement it based on \cite{zhou2018canny} by adding point and line features-based VIO.\\
\textit{\textbf{(iii)} Baseline III \cite{campos2021orb}: A SOTA frame camera-based SLAM method}, which estimates the drone’s pose using monocular images and IMU measurements.
This method is used to assess the benefits of upgrading frame cameras to event cameras and incorporating prior 3D point maps.  

\begin{table}[t]
\centering
\normalsize
\begin{tabular}{cccc}
\hline
EV-Pose & Baseline-I & Baseline-II & Baseline-III \\ \hline
\textbf{10.08 \textit{ms}}   & 23.11 \textit{ms}      & 31.76 \textit{ms}       & 29.27 \textit{ms}        \\ \hline
\end{tabular}
\caption{Latency of different systems.}
\label{latencytable}
\vspace{-0.6cm}
\end{table}

% \noindent
\textbf{Evaluation metrics.}
To thoroughly assess the accuracy of EV-Pose's 6-DoF pose estimation, we employed four quantitative metrics: 
$(i)$ absolute translation error (ATE); 
$(ii)$ absolute rotation error (ARE); 
$(iii)$ relative translation error (RTE); 
$(iv)$ relative rotation error (RRE).
The absolute error measures the accuracy of the drone's pose estimates, while the relative error assesses the consistency of pose estimates over time, both crucial for drone flight control. 
The estimated pose is aligned with the ground truth using the tool \cite{grupp2017evo}.

\begin{figure*}[t]
\setlength{\abovecaptionskip}{-0.05cm} 
\setlength{\belowcaptionskip}{-0.35cm}
\setlength{\subfigcapskip}{-0.2cm}
% \begin{minipage}[b]{0.9\columnwidth}
\centering
    \subfigure[Mean of ATE]{
        \centering
        \includegraphics[width=0.49\columnwidth]{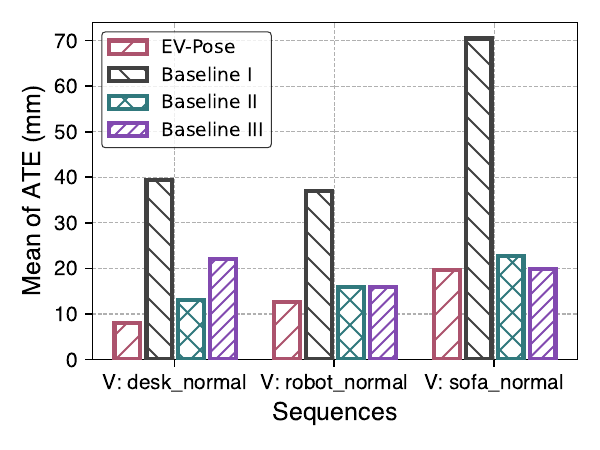}
    }
    \subfigure[Mean of RTE]{
        \centering
        \includegraphics[width=0.49\columnwidth]{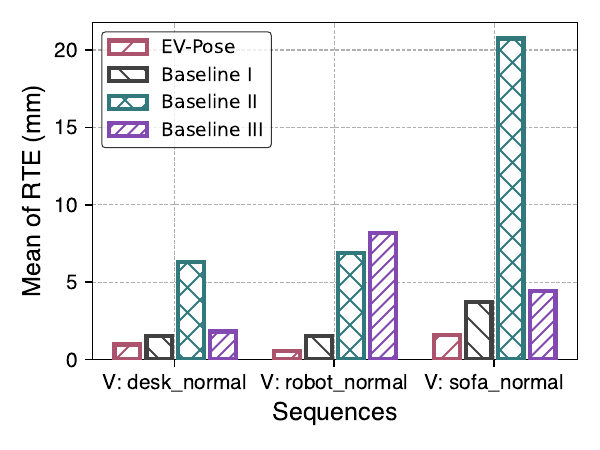}
    }
    \subfigure[Mean of ARE]{
        \centering
        \includegraphics[width=0.48\columnwidth]{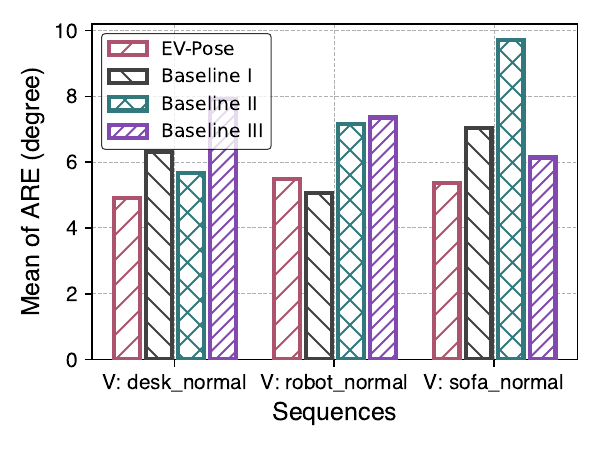}
    }
    \subfigure[Mean of RRE]{
        \centering
        \includegraphics[width=0.48\columnwidth]{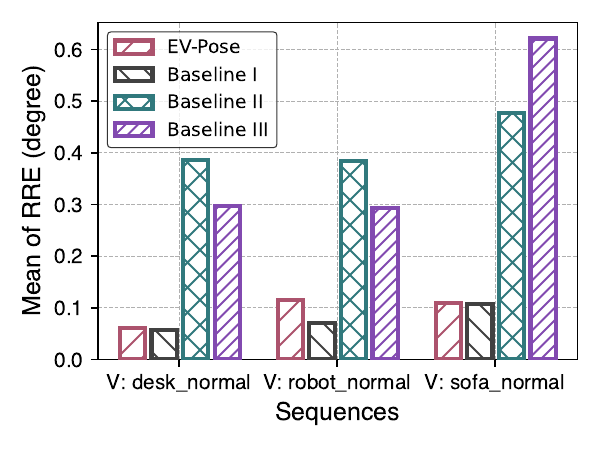}
    }
\caption{Overall performance of \textit{EV-Pose} and three 6-DoF pose tracking systems on VECtor dataset.}
\label{fig:Overall_VEC}
\vspace{-0.1cm}
\end{figure*}

\begin{figure*}[t]
\setlength{\abovecaptionskip}{-0.05cm} 
\setlength{\belowcaptionskip}{-0.35cm}
\setlength{\subfigcapskip}{-0.2cm}
% \begin{minipage}[b]{0.9\columnwidth}
\centering
    \subfigure[Mean of ATE]{
        \centering
        \includegraphics[width=0.49\columnwidth]{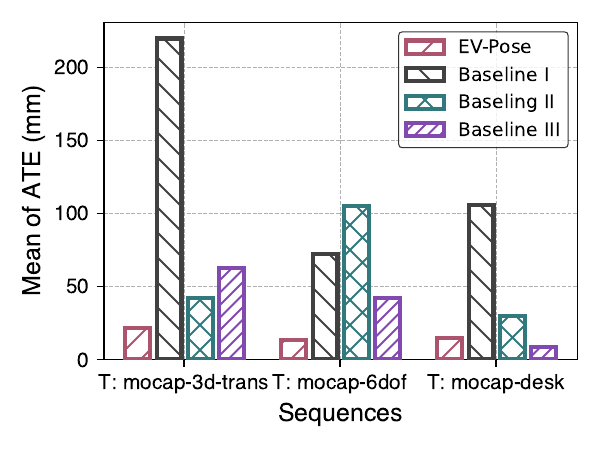}
    }
    \subfigure[Mean of RTE]{
        \centering
        \includegraphics[width=0.49\columnwidth]{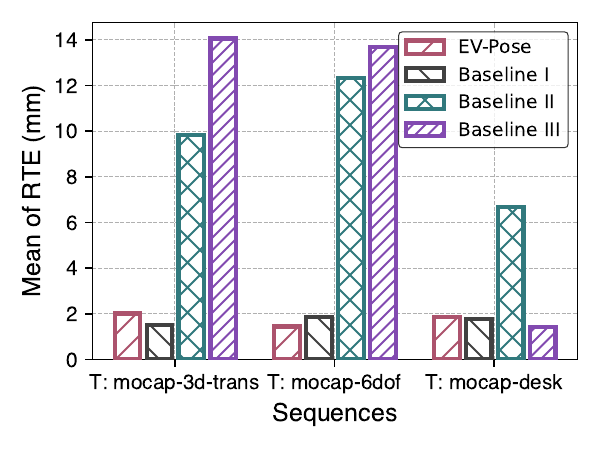}
    }
    \subfigure[Mean of ARE]{
        \centering
        \includegraphics[width=0.48\columnwidth]{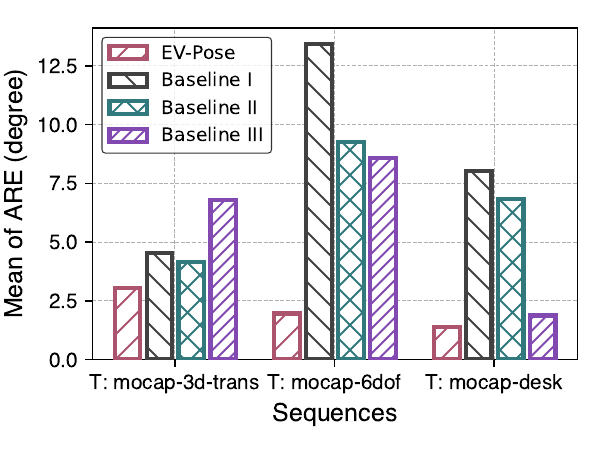}
    }
    \subfigure[Mean of RRE]{
        \centering
        \includegraphics[width=0.48\columnwidth]{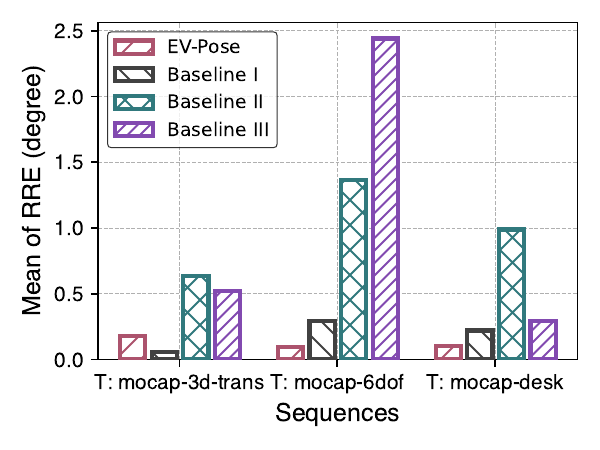}
    }
\caption{Overall performance of EV-Pose and three 6-DoF pose tracking systems on TUM-VIE dataset.}
\label{fig:Overall_TUM}
\vspace{0.1cm}
\end{figure*}

% \noindent 
\textbf{Robustness evaluation.}
To validate robustness of EV-Pose, we conduct field studies in a controlled indoor environment as shown in \fig \ref{setup}(c)-\fig \ref{setup}(f), evaluating its performance under various dynamic conditions, including different drone velocities, frequencies, HDR conditions, and scene dynamics.

% \vspace{-0.3cm}
\subsection{Overall Performance}

% \noindent 
\textbf{Field studies result.} 
We first evaluate EV-Pose's pose tracking performance in an indoor environment of \fig \ref{setup}(a). 
As shown in \fig \ref{fig:overall_indoor}, EV-Pose achieves a mean ATE of 6.9$mm$, outperforming Baseline I, Baseline II, and Baseline III by 77.4\%, 58.1\%, and 66.8\%, respectively. 
For rotation, its mean ARE is 1.34$\degree$, with improvements of 59.3\%, 53.8\%, and 66.0\% over all baselines.
Furthermore, we assess EV-Pose in an outdoor environment of \fig \ref{setup}(b).
As shown in \fig \ref{fig:overall_outdoor}, EV-Pose achieves a mean ATE of 12.6 $mm$, again surpassing the three baselines by 55.3\%, 35.7\%,  and 50.9\%, respectively. 
For rotation, its mean ARE is 1.42 $\degree$, with improvements of 62.2\%, 52.3\%, and 57.0\% over all baselines.
During operation, EV-Pose maintains a latency of 10.08$ms$, compared to Baseline I, II, and III latencies of 23.11$ms$, 31.76$ms$, and 29.27$ms$, as shown in Tab. \ref{latencytable}.
The superior performance of EV-Pose stems from two key factors:
$(i)$ The integration of a 3D point reference map and 2D event-3D point matching significantly enhances pose tracking, outperforming Baseline I and III through the use of external information.
$(ii)$ Motion blur from drone flight impairs edge detection in RGB images, while their low frame rate causes large pose changes in dynamic scenes. In contrast, event cameras with $ms$-level latency, unaffected by motion blur, enable rapid and accurate edge tracking. EV-Pose's high pose tracking frequency leads to smaller pose changes, further enhancing tracking accuracy.

% \noindent 
\textbf{Dataset-based studies result.}
We further evaluate the performance of EV-Pose using public datasets.
\fig \ref{fig:Overall_VEC} and \fig \ref{fig:Overall_TUM} illustrate the overall 6-DoF pose tracking performance of EV-Pose compared to three competitive SOTA methods.
\fig \ref{fig:Overall_VEC} shows EV-Pose's performance on the VECtor dataset, where it consistently achieves superior accuracy across various scenarios.
EV-Pose has a mean ATE of 13.19$mm$, outperforming Baseline I, Baseline II, and Baseline III by 72.63\%, 22.42\%, and 30.62\%, respectively.
EV-Pose's mean of ARE is 1.05$\degree$, outperforming the same methods by 53.33\%, 90.72\%, and 78.13\%.
For RTE, EV-Pose's mean of RTE is 5.25$mm$, surpassing the others by 14.36\%, 30.09\%, and 26.37\%.
Lastly, its mean of RRE is 0.095 $\degree$, beating other methods by up to 77.14\%.
\fig \ref{fig:Overall_TUM} presents results on the TUM-VIE dataset, showing similar trends. The mean of ATE is 16.59$mm$, ARE is 2.12$\degree$, RTE is 1.78$mm$, and RRE is 0.12$\degree$, with EV-Pose outperforming Baseline I, II, and III in each metric. 
EV-Pose excels in challenging environments with dynamic changes, fast motions, and HDR conditions.
By upgrading frame camera to event camera, 
EV-Pose consistently demonstrates resilience across various conditions and achieves lower errors. 
This underscores its robustness and reliability as a solution for real-world 6-DoF pose tracking applications.

\begin{figure*}
\setlength{\abovecaptionskip}{-0.05cm} 
\setlength{\belowcaptionskip}{-0.35cm}
\setlength{\subfigcapskip}{-0.2cm}
    \begin{minipage}[b]{1.03\columnwidth}
        \subfigure[Absolute estimation error]{
            \centering
            \includegraphics[width=0.46\columnwidth]{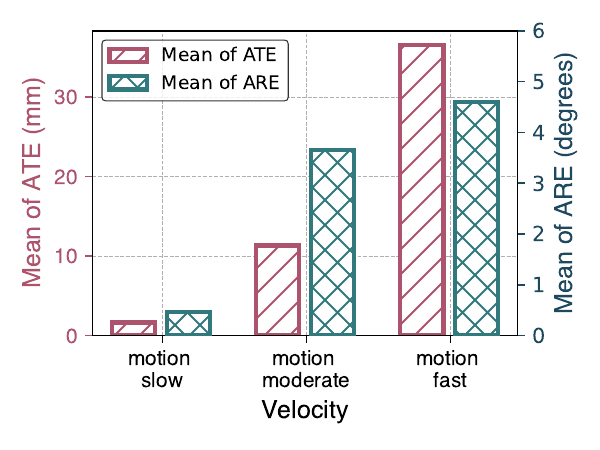}
        }
        \subfigure[Relative estimation error]{
            \centering
            \includegraphics[width=0.46\columnwidth]{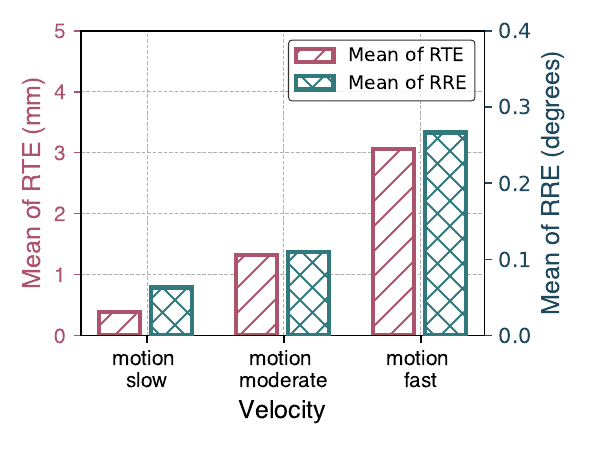}
        }
        \caption{Impact of drone moving velocity.}
        \label{fig:robust_velocity}
    \end{minipage}
    \begin{minipage}[b]{1.03\columnwidth}
        \subfigure[Absolute estimation error]{
            \centering
            \includegraphics[width=0.46\columnwidth]{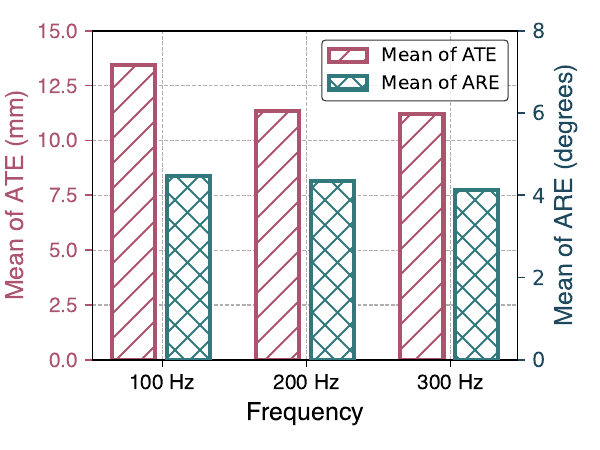}
        }
        \subfigure[Relative estimation error]{
            \centering
            \includegraphics[width=0.46\columnwidth]{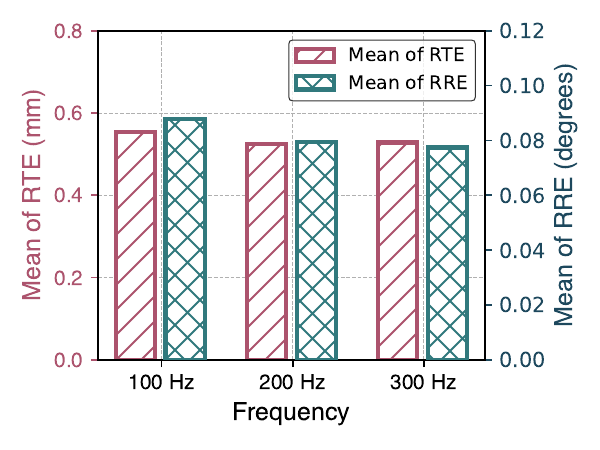}
        }
        \caption{Impact of estimation frequency.}
        \label{fig:robust_freq}
    \end{minipage}
\vspace{-0.2cm}
\end{figure*}

\begin{figure*}
\setlength{\abovecaptionskip}{-0.05cm} 
\setlength{\belowcaptionskip}{-0.35cm}
\setlength{\subfigcapskip}{-0.2cm}
    \begin{minipage}[b]{1.03\columnwidth}
        \subfigure[Absolute estimation error]{
            \centering
            \includegraphics[width=0.46\columnwidth]{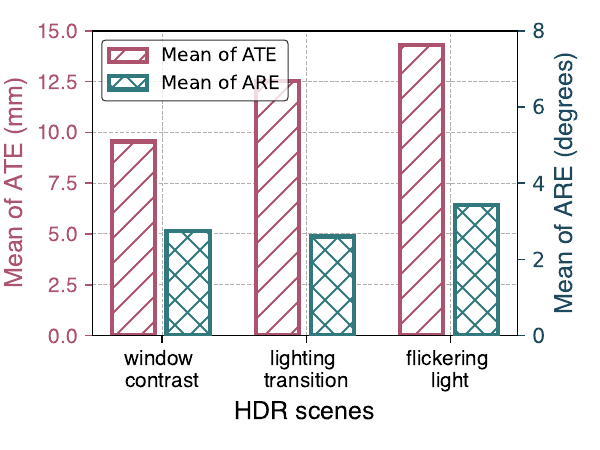}
        }
        \subfigure[Relative estimation error]{
            \centering
            \includegraphics[width=0.46\columnwidth]{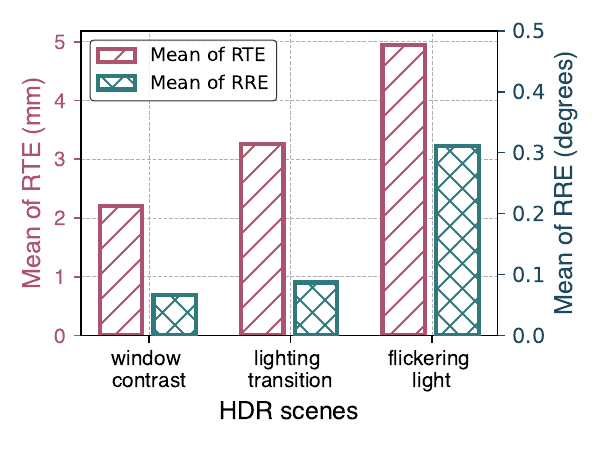}
        }
        \caption{Impact of HDR.}
        \label{fig:robust_HDR}
    \end{minipage}
    \begin{minipage}[b]{1.03\columnwidth}
        \subfigure[Absolute estimation error]{
            \centering
            \includegraphics[width=0.46\columnwidth]{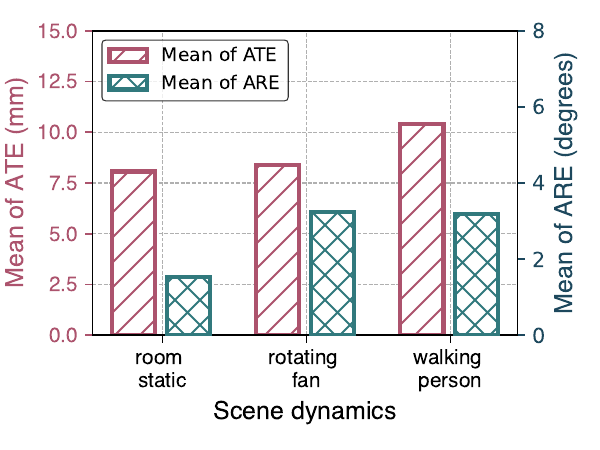}
        }
        \subfigure[Relative estimation error]{
            \centering
            \includegraphics[width=0.46\columnwidth]{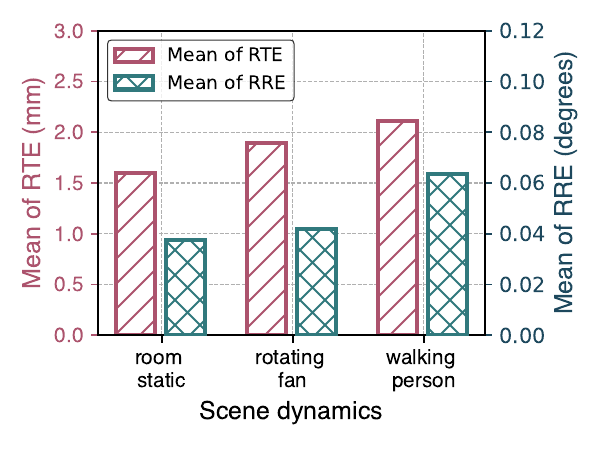}
        }
        \caption{Impact of scene dynamics.}
        \label{fig:robust_dyna}
    \end{minipage}
\vspace{-0.2cm}
\end{figure*}

% \vspace{-0.4cm}
\subsection{Robustness Evaluation}
Our robustness experiments primarily focus on evaluating EV-Pose's performance under extreme conditions.

% \noindent 
\textbf{Impact of drone moving velocity.}
To evaluate the impact of drone velocity on EV-Pose, we conduct experiments at various speeds. 
As shown in \fig \ref{fig:robust_velocity}, we separately measure both absolute and relative errors across different translation and rotation velocities.  
For translation velocity, we categorize motion into slow (\(v < 2\) m/s), moderate (\(2\) m/s \(< v < 4\) m/s), and fast (\( v > 4\) m/s) to assess performance. 
% (\fig \ref{fig:robust_velocity}(a)). 
The mean ATE of EV-Pose is 1.6 $mm$, 11.3 $mm$, and 36.7 $mm$ for slow, moderate, and fast motion, respectively. 
% The mean ATE of EV-Pose is 1.6 $mm$, 11.3 $mm$, and 68.8 $mm$ for slow, moderate, and fast motion, respectively. 
Regarding relative errors, the mean RTE is 0.38 $mm$, 1.32 $mm$, and 3.06 $mm$ for corresponding speed categories.  
% Regarding relative errors, the mean RTE is 0.3 $mm$, 0.5 $mm$, and 6.5 $mm$ for corresponding speed categories.  
Similarly, for rotation velocity, we classify motion into slow (\(\omega < 10^\circ\)), moderate (\(10^\circ < \omega < 30^\circ\)), and fast (\( \omega > 30^\circ\)).  
The mean ARE is 0.46°, 3.64°, and 4.58°, while mean RRE is 0.06°, 0.109°, and 0.26°.  
These results highlight the robustness of EV-Pose across different velocities, thanks to its integration of an event camera with a 3D point map, which enables precise tracking at high speeds without being affected by motion blur.

% \noindent 
\textbf{Impact of estimation frequency.}
We examine the impact of estimation frequency on the performance of EV-Pose, as shown in \fig \ref{fig:robust_freq}. 
Our experiments include varying frequencies: 100$Hz$, 200$Hz$, and 300$Hz$.
As the frequency increased from 100$Hz$ to 300$Hz$, the mean of ATE decreased from 13.4$mm$ to 11.2$mm$, the mean of ARE from 4.46$\degree$ to 4.13$\degree$, the mean of RTE from 0.55$mm$ to 0.52$mm$, and the mean of RRE from 0.087$\degree$ to 0.077$\degree$. 
Higher frequencies result in smaller changes between drone poses, providing EV-Pose with more information for accurate pose estimation. 
The results demonstrate the robustness of EV-Pose to varying frequencies, making it a promising solution for high-frequency pose estimation in fast-moving drones.

% \noindent 
\textbf{Impact of HDR.}
We investigate the influence of HDR on the performance of EV-Pose. Various HDR scenarios are triggered:
$(i)$ \textit{window contrast}: luminance differences between different areas (\fig \ref{setup}(c));
$(ii)$ \textit{lighting transition}: switching from lights-on to lights-off in area of \fig \ref{setup}(a); 
$(iii)$ \textit{flickering light}: the flickering light generated by rapidly switching a controllable spotlight (\fig \ref{setup}(d)).
As shown in \fig \ref{fig:robust_HDR}, EV-Pose’s mean of ATE, ARE under different scenarios are 9.5$mm$, 12.5$mm$, 14.3$mm$; 2.74$\degree$, 2.60$\degree$, 3.43$\degree$; mean of RTE, and RRE are 2.2$mm$, 3.26$mm$, 4.9$mm$; and 0.067$\degree$, 0.087$\degree$, 0.31$\degree$, respectively. 
EV-Pose demonstrates stability under HDR conditions, highlighting its potential for reliable performance in real-world environments with varying lighting conditions.

% \noindent
\textbf{Impact of scene dynamics.}
We finally assess the impact of scene dynamics on performance. Our setup included:
$(i)$ \textit{room static}: a completely static room (\fig \ref{setup}(c));
$(ii)$ \textit{rotating fan}: a running fan produces numerous events, moderately rendering scene modeling outdated (\fig \ref{setup}(e)); 
$(iii)$ \textit{walking people}: several people walking across the test area (\fig \ref{setup}(f)). 
Results are shown in \fig \ref{fig:robust_dyna}.
The mean of ATE for EV-Pose in each scenario is 8.1$mm$, 8.4$mm$, and 10.4$mm$, respectively. The mean of ARE is 1.53$\degree$, 3.24$\degree$, and 3.19$\degree$. The mean of RTE is 1.6$mm$, 1.9$mm$, and 2.1$mm$. The mean of RRE is 0.037$\degree$, 0.042$\degree$, and 0.063$\degree$. 
EV-Pose demonstrates resilience against these challenges. 
By combining the drone’s motion information as \textit{proprioceptive} measurements with 3D point map–based \textit{exteroceptive} measurements, EV-Pose effectively isolates and removes the impact of dynamic objects. This ensures accurate and consistent results under varying scene dynamics and mitigates issues caused by outdated scene models, strengthening EV-Pose’s reliability in dynamic environments.

\begin{figure*}
\setlength{\abovecaptionskip}{-0.05cm}  
\setlength{\belowcaptionskip}{-0.35cm}
\setlength{\subfigcapskip}{-0.2cm}
    \begin{minipage}[b]{1\columnwidth}
        \subfigure[CDF of ATE]{
            \centering
            \includegraphics[width=0.46\columnwidth]{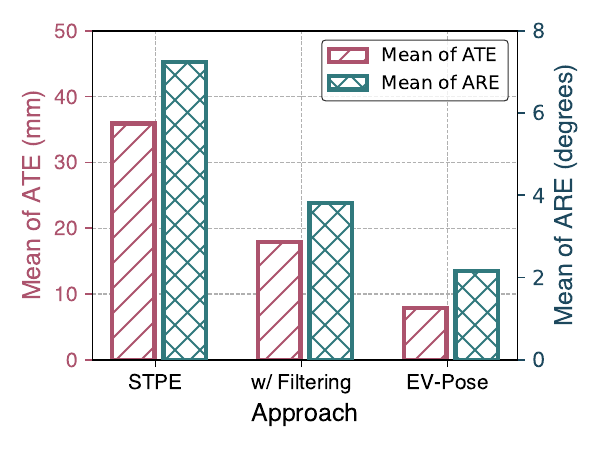}
        }
        \subfigure[Latency analysis]{
            \centering
            \includegraphics[width=0.46\columnwidth]{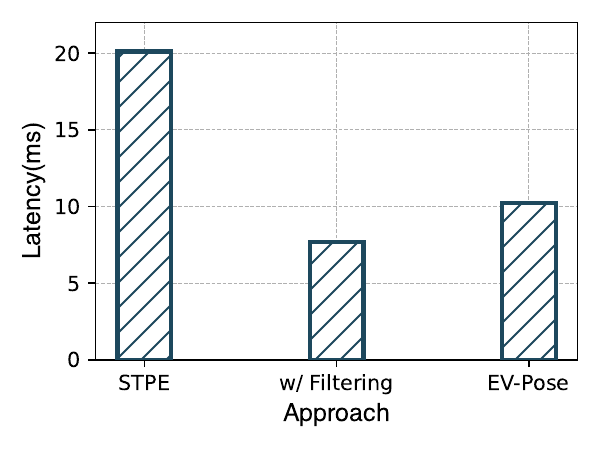}
        }
        \caption{Impact of modules.}
        \label{fig:ablation_module}
    \end{minipage}
    \begin{minipage}[b]{1\columnwidth}
        \subfigure[CDF of ATE]{
            \centering
            \includegraphics[width=0.46\columnwidth]{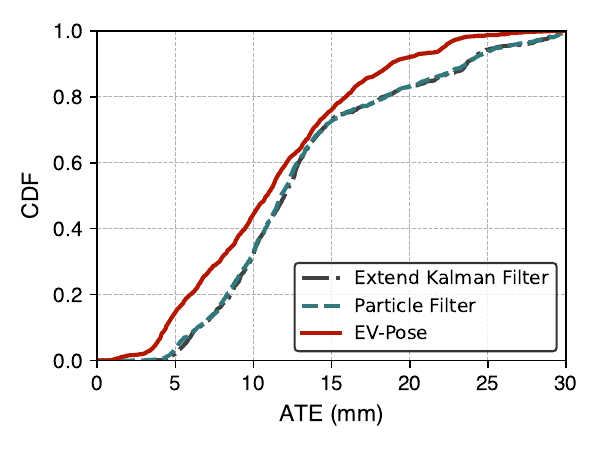}
        }
        \subfigure[CDF of ARE]{
            \centering
            \includegraphics[width=0.46\columnwidth]{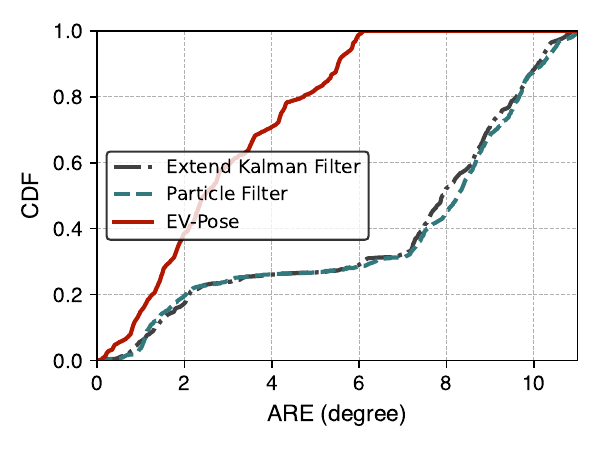}
        }
        \caption{Comparison of fusion framework.}
        \label{fig:abalation_opti}
    \end{minipage}
\vspace{-0.1cm}
\end{figure*}

\begin{figure*}
\setlength{\abovecaptionskip}{-0.05cm} 
\setlength{\belowcaptionskip}{-0.35cm}
\setlength{\subfigcapskip}{-0.2cm}
  \begin{minipage}[t]{0.66\columnwidth}
    \centering
    \includegraphics[width=1\columnwidth]{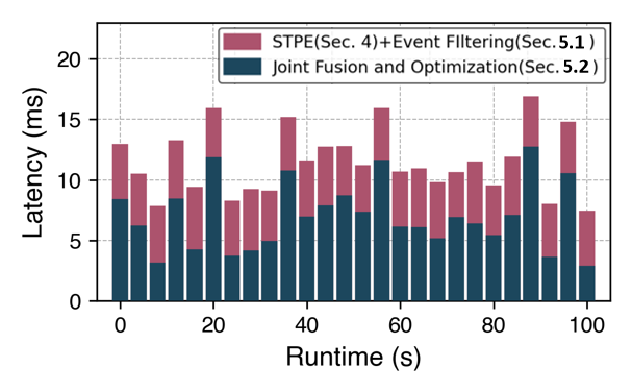}
    \caption{System latency.}
    \label{fig:latency}
  \end{minipage}
  \begin{minipage}[t]{0.67\columnwidth}
    \centering
    \includegraphics[width=1\columnwidth]{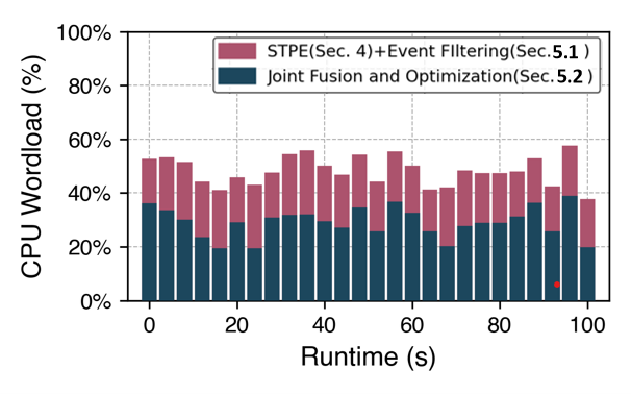}
    \caption{CPU workload.}
    \label{fig:cpu}
  \end{minipage}
  \begin{minipage}[t]{0.66\columnwidth}
    \centering
    \includegraphics[width=1\columnwidth]{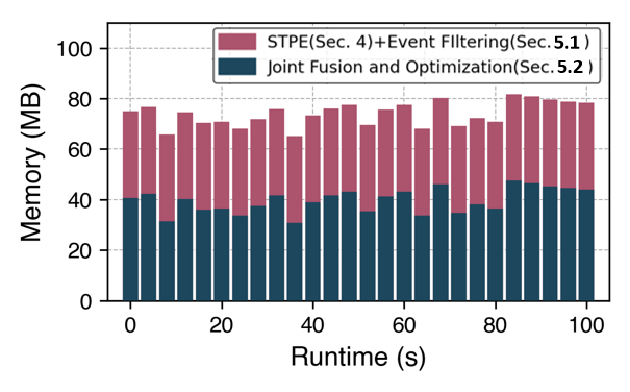}
    \caption{Memory usage.}
    \label{fig:memory}
  \end{minipage}
  \hfill
  \vspace{-0.1cm}
\end{figure*}

% \vspace{-0.3cm}
\subsection{Micro Benchmarks}

\textbf{Effectiveness of each module.}
To evaluate how each module contributes to EV-Pose, we incrementally added fusion modules to the event-only baseline STPE (§ \ref{4}). 
The experiment configurations are set as 
$(i)$ \textit{STPE}: The event-only pose update method.  
$(ii)$ \textit{w/ motion-instructed filtering}: STPE with early-stage fusion to filter out noisy events. 
$(iii)$ \textit{EV-Pose}: The proposed approach applies hierarchical fusion and optimization, including both early-stage and later-stage fusion.
As shown in \fig \ref{fig:ablation_module}(a), the baseline mean ATE and ARE are 35.9$mm$ and 7.12$\degree$.
Adding Event Filtering (§ \ref{5.1}) module, these values decreased to 24.2$mm$ and 6.67$\degree$. 
With the motion-aware hierarchical fusion and optimization (§ \ref{5.3}), EV-Pose achieves ATE at 6.9$mm$ and ARE at 1.34$\degree$, creating the best performance. 
Also, latency comparison is shown in \fig \ref{fig:ablation_module}(b).
Applying event filtering, latency decreases from 20.11$ms$ to 7.69$ms$, which shows the effectiveness of filter modules.
After performing the factor graph optimization, the latency slightly increased to 10.08$ms$, indicating that integrating the IMU into the factor graph effectively reduces absolute error while introducing only minimal and acceptable latency.

% \noindent 
\textbf{Effectiveness of event filtering.}
As illustrated in \fig \ref{fig:ablation_module}, our early-stage fusion of event filtering significantly enhances effectiveness of distance field map production, thereby improving overall accuracy and latency performance. 
Specifically, more than \textit{40\%} of extraneous and irrelevant events are filtered out, leading to higher accuracy and lower latency. 

\textbf{Comparison of fusion frameworks.}
We compared our factor graph-based fusion framework with two other methods: particle filter (PF) and extended Kalman filter (EKF). 
As shown in \fig \ref{fig:abalation_opti}, EV-Pose improves translation tracking by over 25.41\% and 25.7\% compared to PF and EKF, and rotation tracking by over 57.81\% and 58.43\%. 
This performance is attributed to the MHFO module's ability to tightly integrate exteroceptive and proprioceptive measurements.

\textbf{Latency analysis.}
As shown in \fig \ref{fig:latency}, we measure the end-to-end latency of EV-Pose, including delays from the \textit{STPE + Event Filtering} and \textit{Joint Fusion and Optimization} modules. 
The average latency is 10.08$ms$, with \textit{STPE  + Event Filtering} contributing 3.05$ms$ and \textit{Joint Fusion and Optimization} 7.03$ms$, which is suitable for use in flight control loops.

% \noindent 
\textbf{Resource Overhead.}
\fig \ref{fig:cpu} and \fig \ref{fig:memory} show that EV-Pose's CPU usage remains $<$60\%, with average memory usage $<80 MB$, including $7 MB$ for storing point maps.
These results highlight EV-Pose's minimal resource overhead, making it suitable for resource-constrained drone systems.

%% file: relatework.tex
% \vspace{-0.5cm}
\section{Related Work}\label{7}
% \noindent 
\textbf{6-DoF pose tracking of drone.}
As a key enabler of drone applications (\eg, instant delivery), numerous 6-DoF pose tracking systems have been proposed over the past decade. 
Most existing solutions leverage onboard sensors (\eg, IMU, RGB/D cameras, LiDAR, and mmWave radar) for drone pose tracking.
IMU-based methods infer the motion of a drone from built-in gyroscope and accelerometer, though cost-effective, suffer from severe cumulative drift \cite{Zhao2024foes, brossard2020ai}. 
Vision-based methods use RGB/D cameras to track relative poses via feature matching \cite{cao2022edge, chi2021locate}. 
Recent researches integrate visual odometry with IMU for better performance \cite{zhou2020ego, qin2018online}.
Modern VPS for drone 6-DoF pose tracking further combines scene maps with 2D-3D matching for global pose estimation \cite{hu2023efficient, mur2017visual}.
However, these solutions face limitations of 
$(i)$ Motion blur and large pose changes between frames may lead to tracking failures;
$(ii)$ VIO operates at <30 Hz, while 2D-3D matching is resource intensive, limiting map-based tracking to <1 Hz \cite{hu2023efficient, pang2023ubipose}, both falling short of the flight controller's requirements.
Previous studies also explored the use of LiDAR \cite{vrf, recap}, mmWave radar \cite{sie2023batmobility}. 
However, limitations in accuracy and latency hinder their widespread adoption \cite{vidal2018ultimate}.

Compared to previous methods, EV-Pose is the first to redesign a drone-oriented VPS using an event camera with $ms$-level latency, enabling accurate, low-latency drone pose tracking.
In addition, its sensor setup offers strong resistance to lighting changes, scene dynamics, and outdated maps, thanks to the high dynamic range of event cameras and the combined use of exteroceptive and proprioceptive measurements to reduce external environmental impact.

\textbf{Event-based 6-DoF pose tracking.}
Leveraging the low-latency advantages of an event camera for 6-DoF pose tracking can improve system robustness.
Current methods rely on contrast maximization with frame-to-frame warping or learning-based approaches, but they are limited to homography transformations and need extensive training data, lacking generality guarantees \cite{liu2020globally, peng2021globally}.
Several methods combine event cameras with IMUs, using extended Kalman filters or continuous time representations for 6-DoF pose estimation \cite{mueggler2018continuous}, or fuse event data with RGB camera brightness information to track features \cite{cao2024EventBoost}. 
They suffer from data noise and fail to fully utilize event cameras due to heavy reliance on RGB, causing accuracy and latency bottlenecks.
\cite{wang2025ultra} performs drone ground localization by fusing an event camera with an mmWave radar, while \cite{xu2023taming, falanga2020dynamic} use dual event cameras and an IMU for external obstacle localization.
However, none of these methods provides drone 6-DoF ego-motion estimation. 

To address accuracy and latency bottlenecks, EV-Pose is the first to incorporate a prior 3D point cloud to improve event camera–based 6-DoF drone pose estimation.
EV-Pose introduces an STPE module to extract a temporal distance field, enabling cross-modality matching for pose estimation. 
Then, by incorporating motion information, EV-Pose mitigates event bursts and fully exploits event cameras for accurate and efficient 6-DoF drone pose tracking.

%% file: conclusion.tex
% \vspace{-0.3cm}
\section{Conclusion} \label{8}

EV-Pose redesigns drone-oriented VPS for accurate and efficient drone landing using the event camera with $ms$-level latency. 
Leveraging temporal relationships among events, EV-Pose designs the STPE module to extract a temporal distance field and model a 2D event-3D point map matching problem for pose estimation. 
Incorporating motion data, EV-Pose then proposes the MHFO scheme to hierarchically fuse event and IMU to enhance the accuracy and efficiency of matching.
Experiments show its superior performance, highlighting its potential for drone-based applications.